\documentclass[12pt]{article}

\usepackage{tabularx}
\usepackage{wrapfig}
\usepackage{titlesec}
\usepackage{mathpazo}
\usepackage{ebgaramond}
\usepackage{charter}
\usepackage{bbold}
\usepackage{booktabs}
\usepackage{array}
\usepackage{multirow}
\usepackage{pifont}
\usepackage{xcolor}
\usepackage[cal=boondoxo]{mathalfa}
\usepackage{bm}
\usepackage{twemojis}
\definecolor{myblue}{RGB}{13, 71, 161}   %
\definecolor{myred}{RGB}{232, 93, 117}   %
\usepackage{tikz}
\usetikzlibrary{positioning, arrows.meta}
\usetikzlibrary{calc} 
\usepackage{pdflscape}
\usepackage{adjustbox}
\usepackage{subcaption}
\usepackage{colortbl} %
\usepackage{longtable} %
\renewcommand{\arraystretch}{1.2}

\definecolor{customgray}{gray}{10.5}  %
\newcommand{\mycomment}[1]{}
\titleformat{\section}
{\normalfont\Large\bfseries\color{customgray}} %
{\thesection}{1em}{}

\titleformat{\subsection}
{\normalfont\large\bfseries\color{gray}} %
{\thesubsection}{1em}{}
\titleformat{\subsubsection}
{\color{gray}\normalsize\bfseries} %
{\thesubsubsection}{1em}{}
\titlespacing*{\section}{0pt}{1.5ex plus 1ex minus .2ex}{1.5ex plus .2ex}
\titlespacing*{\subsection}{0pt}{1.25ex plus 1ex minus .2ex}{1.25ex plus .2ex}
\usepackage{bm}
\usepackage[most]{tcolorbox}
\newtcolorbox{examplebox}{
  enhanced,
  breakable,
  colback=blue!5,
  colframe=blue!60!black,
  boxrule=0.8pt,
  arc=2mm,
  left=4mm,
  right=4mm,
  top=2mm,
  bottom=2mm
}
\newtcolorbox{titledexamplebox}[1]{
  enhanced,
  breakable,
  colback=blue!5,
  colframe=blue!60!black,
  title=\textbf{#1},
  fonttitle=\bfseries,
  coltitle=white,
  boxrule=0.3pt,
  arc=0mm
}
\newtcolorbox{exampleboxthree}[1]{
  enhanced,
  breakable,
  colback=white,
  colframe=myred!20,
  boxrule=0.5pt,
  leftrule=3pt,
  colbacktitle=white,
  coltitle=myred!80!black,
  fonttitle=\bfseries\sffamily,
  attach boxed title to top left={yshift=-2mm, xshift=4mm},
  boxed title style={boxrule=0pt, colframe=white, colback=white},
  title={#1},
  top=5mm, bottom=5mm, left=5mm, right=5mm,
  enlarge top at break by=6mm,    %
  enlarge bottom at break by=6mm, %
  bottomrule at break=0pt,
  toprule at break=0pt
}
\usepackage{FiraSans}
\newtcolorbox{exampleboxtwo}[1]{
  enhanced,
  breakable,
  colback=myblue!2,
  colframe=myblue!20,
  boxrule=0.5pt,
  leftrule=3pt,
  colbacktitle=white,
  coltitle=myblue!80!black,
  fonttitle=\bfseries\sffamily,
  attach boxed title to top left={yshift=-2mm, xshift=4mm},
  boxed title style={boxrule=0pt, colframe=white, colback=white},
  title={#1},
  top=5mm, bottom=5mm, left=5mm, right=5mm,
  enlarge top at break by=6mm,
  enlarge bottom at break by=6mm,
  bottomrule at break=0pt,
  toprule at break=0pt
}

\usepackage[isbn=false,url=false]{biblatex}
\usepackage{times}
\usepackage{tabularray}
\usepackage{heuristica}
\usepackage{wrapfig}
\usepackage{multirow}
\usepackage{booktabs}
\usepackage{algorithm}
\usepackage{algpseudocode}
\usepackage{makecell}

\usepackage[hmargin=1cm]{geometry}

\usepackage{tikz}

\usetikzlibrary{backgrounds}
\usetikzlibrary{patterns}
\usetikzlibrary{bayesnet}
\usetikzlibrary{arrows}
\usepackage{color}

\usetikzlibrary{backgrounds}

\usepackage{array}

\usepackage{amsmath,amsfonts}
\usepackage{amssymb} %
\usepackage{mathtools}

\usepackage{listings}
\usepackage[most]{tcolorbox}

\definecolor{usercolor}{RGB}{255, 108, 108} %
\definecolor{assistantcolor}{RGB}{255, 189, 69} %

\newtcolorbox{userbox}{
    colback=usercolor!5!white,
    colframe=usercolor!75!black,
    width=\linewidth-1cm,
    left=0.5cm,
    right=0.5cm,
    boxrule=0.4mm,
    arc=2mm,
    fonttitle=\bfseries,
    title=Prompt,
    breakable
}
\usepackage{comment}
\usepackage{booktabs}

\newtcolorbox{assistantbox}{
    colback=assistantcolor!5!white,
    colframe=assistantcolor!75!black,
    width=\linewidth-1cm,
    right=0.5cm,
    left=0.5cm,
    boxrule=0.4mm,
    arc=2mm,
    fonttitle=\bfseries,
    title=Assistant,
    breakable
}

\lstdefinestyle{pythonstyle}{
    language=Python,
    basicstyle=\ttfamily\small\color{black},           %
    keywordstyle=\color{teal}\bfseries,                 %
    stringstyle=\color{brown},                          %
    commentstyle=\color{gray},                          %
    numberstyle=\tiny\color{gray},                      %
    frame=single,                                       %
    backgroundcolor=\color{gray!5},                     %
    breaklines=true,                                    %
    showstringspaces=false,                             %
    numbers=left,                                       %
    xleftmargin=1em,                                    %
    rulecolor=\color{black},  
    morecomment=[s][\color{olive!80!black!80}\slshape]{"""}{"""},  %
}

\lstdefinestyle{bashstyle}{
    language=bash,
    basicstyle=\ttfamily\small,
    frame=single,
    backgroundcolor=\color{gray!10},
    breaklines=true,
    showstringspaces=false
}

\usepackage{amsmath}
\usepackage{amssymb}
\usepackage{mathtools}
\usepackage{amsthm}
\theoremstyle{plain}

\theoremstyle{definition}

\theoremstyle{remark}

\definecolor{bronze}{RGB}{205, 127, 50}
\definecolor{silver}{RGB}{192, 192, 192}
\definecolor{gold}{RGB}{255, 215, 0}
\usepackage{graphicx}   %
\usepackage{amsmath}    %
\usepackage{amssymb}    %
\usepackage{fancyhdr}   %
\usepackage{geometry}   %
\usepackage{titlesec}   %
\usepackage{xcolor}     %
\usepackage{caption}  %
\titleformat{\section}{\normalfont\bfseries\Large}{\thesection}{1em}{}
\titlespacing*{\section}{0pt}{*3}{*2}

\newtoggle{final}
\togglefalse{final} %

\newcolumntype{C}{>{\centering\let\newline\\\arraybackslash\hspace{0pt}}m{2cm}}

\newcommand\vartextvisiblespace[1][.5em]{%
  \makebox[#1]{%
    \kern.07em
    \vrule height.3ex
    \hrulefill
    \vrule height.3ex
    \kern.07em
  }%
}

\usepackage{tcolorbox}

\usepackage{url}      %
\usepackage{hyperref}   %

\definecolor{headerblue}{RGB}{250, 245, 245}
\definecolor{titleblack}{RGB}{80, 0, 0} %
\definecolor{abstractblack}{RGB}{80, 0, 0} %

\newtcolorbox{fullbox}{
colback=headerblue,
colframe=white,
width=\textwidth,
boxrule=0pt,
arc=10pt,
outer arc=10pt,
boxsep=10pt,
left=10pt,
right=10pt,
top=10pt,
bottom=10pt
}

\author
{Firstname Lastname,$^{1}$ 
\\
\normalsize{$^{1}$Huawei Noah’s Ark Lab, London, UK.}\\
\normalsize{$^{2}$Technical University of Darmstadt, Darmstadt, Germany.}\\
\normalsize{$^{3}$University College London, London, UK.}\\
\\
\normalsize{
Correspondence: 
firstname.lastname@huawei.com,} \\
}

\date{}

\begin{document}

\begin{fullbox}
\vspace{0.1em}
\begin{center}
{\fontsize{17}{20}\selectfont \bfseries\sffamily
Decoding as Optimisation on the Probability Simplex:
\par}
\vspace{0.2em}
{\fontsize{12.5}{15}\selectfont \sffamily\color{black!60}
From Top-K to Top-P (Nucleus) to Best-of-K Samplers
\par}

\vspace{1em}
\textbf{\textsf{Xiaotong Ji$^{1,\dagger}$, Rasul Tutunov$^{1}$, Matthieu Zimmer$^{1}$}, Haitham Bou-Ammar$^{1,2, \dagger}$}\\
{\small $^1$ Huawei Noah's Ark} {\small $\ ^2$ AI Centre, Department of Computer Science, UCL}\\
{\small$\dagger$ Equal contributions}

\end{center}

\noindent

\textbf{Abstract:} 
Decoding sits between a language model and everything we do with it, yet it is still treated as a heuristic knob-tuning exercise. We argue decoding should be understood as a principled optimisation layer: at each token, we solve a regularised problem over the probability simplex that trades off model score against structural preferences and constraints. This single template recovers greedy decoding, Softmax sampling, Top-K, Top-P, and Sparsemax-style sparsity as special cases, and explains their common structure through optimality conditions.

More importantly, the framework makes it easy to invent new decoders without folklore. We demonstrate this by designing Best-of-K (BoK), a KL-anchored coverage objective aimed at multi-sample pipelines (self-consistency, reranking, verifier selection). BoK targets the probability of covering good alternatives within a fixed K-sample budget and improves empirical performance. We show that such samples can improve accuracy by, for example, +18.6\% for Qwen2.5-Math-7B on MATH500 at high sampling temperatures. 
\vspace{0.55em}
\begin{tcolorbox}[
  enhanced,
  boxrule=0pt,
  colback=black!1,
  borderline west={2pt}{0pt}{black!15},
  arc=2mm,
  left=3mm,right=3mm,top=1.0mm,bottom=1.0mm,
  width=\linewidth
]
\centering
{\fontsize{11.2}{13.2}\selectfont\sffamily\bfseries\color{black!90}
\raisebox{0.05ex}{\large \twemoji{pushpin}}\hspace{0.35em}
Our message is simple: \textbf{Decoding is not a hack}; it is optimisation!
}
\end{tcolorbox}
\end{fullbox}

\thispagestyle{fancy}
\section{Introduction}
Large language model (LLM) decoding is usually taught like a cookbook: Top-K \cite{fan2018hierarchicalneuralstorygeneration, noarov2025foundationstopkdecodinglanguage}, temperature \cite{guo2017calibrationmodernneuralnetworks, ji2026scalablepowersamplingunlocking}, Top-P \cite{holtzman2020curiouscaseneuraltext, nguyen2025turningheatminpsampling}, greedy \cite{NIPS2014_5a18e133, shi2024thoroughexaminationdecodingmethods}, beam search \cite{vijayakumar2018diversebeamsearchdecoding, franceschelli2024creativebeamsearchllmasajudge}; a shelf of ``tricks'' you pick from depending on the application\cite{ji2025surelysafealignmentlarge, 10731312, liu2026functionalcorrectnessexploringhallucinations, mitchener2025bixbenchcomprehensivebenchmarkllmbased, tutunov2025modelbasedsampleefficientaiassistedmath, tutunov2024largelanguagemodelsgenerate, woodruff2026acceleratingscientificresearchgemini,zimmer2025bourbakiselfgeneratedgoalconditionedmdps}, whether you want more determinism \cite{gond2026llm42enablingdeterminismllm, suresh2025beaverefficientdeterministicllm}, more diversity \cite{vilnis2023arithmeticsamplingparalleldiverse, wang2025effectsamplingdiversityscaling, zhang2025verbalizedsamplingmitigatemode, zhou2025balancingdiversityriskllm}, or fewer hallucinations \cite{chuang2024doladecodingcontrastinglayers, wang2025mllmseedynamiccorrection, wei2024measuringreducingllmhallucination}. The problem is that this framing makes decoding feel like a bag of heuristics: useful, but conceptually disconnected from the rest of machine learning.

This paper argues the opposite: many decoding strategies are not heuristics at all. They are solutions to explicit optimisation problems, often the same optimisation problem, with different regularisers and constraints. Once you see decoding through that lens, familiar algorithms stop looking like folklore and start looking like principled design choices. Greedy decoding becomes a limiting case of an objective with no regularisation. Softmax sampling becomes the unique optimum of a score-maximisation problem regularised by (negative) Shannon entropy. Sparsity-inducing decoders arise from alternative convex penalties. In short: decoders differ less by ``how they sample'' and more by ``what objective they are implicitly optimising.'' 

Our starting point is a small, but surprisingly powerful, shift in perspective. A decoder does not have to immediately choose a token. At each step, it can first choose a distribution over tokens, and only then sample (or take the mode). This turns decoding into a clean optimisation problem over the probability simplex: pick a distribution that (i) puts mass on high-scoring tokens while (ii) satisfying desirable structural preferences such as smoothness, sparsity, or staying close to a reference distribution. This ``distribution-first'' view is general enough to cover deterministic and stochastic decoding in one line, and it exposes what decoding algorithms do: they are trading off score against regularisation under simplex constraints. 

With that formulation in place, we do something that decoder discussions rarely do: we derive the optimality conditions carefully. Because the variable is a probability distribution, constraints are not decorative; they shape the solution. The simplex geometry introduces the familiar ``active vs inactive'' behaviour: tokens assigned a nonzero probability satisfy an equality condition, while tokens pushed to zero satisfy an inequality. These KKT-style conditions act like a master key: once derived, you can plug in a choice of regulariser and immediately recover the structure of the decoder it implies.

We then use this master key to re-derive canonical decoding rules as special cases. Setting the regulariser weight to zero collapses the optimisation to a linear objective on the simplex, recovering greedy decoding. Choosing negative entropy yields a smooth interior optimum and produces the softmax distribution as a closed-form solution, where temperature sampling is not a hack; it is the optimiser of a maximum-entropy-regularised objective. More broadly, this paper builds the intuition that “decoding methods” are best understood as regularisation families, rather than as unrelated procedures. The takeaway is simple: decoding is an optimisation layer sitting on top of the model’s scores. More precisely, decoding is a convex optimisation problem whose geometry is determined by the choice of regulariser. Once we treat it that way, we gain a principled vocabulary for designing new decoders: decide the behaviour (diversity, sparsity, conservatism, stability), encode it as a regulariser or constraint, and let the resulting optimisation problem define the algorithm. %

Beyond theoretical unification, our framework serves as a generative tool for designing next-generation decoding objectives. We exemplify this by addressing the inefficiencies in current multi-sample pipelines through the design of Best-of-K (BoK). Unlike traditional methods that rely on empirical folklore, BoK utilises a KL-anchored coverage objective to maximise the likelihood of capturing high-quality candidates within a constrained K-sample budget. Our results demonstrate that by optimising for coverage, BoK not only provides a more principled decoding path but also achieves improved practical performance. 

Across two 7B Qwen models (a math-specialised variant and a general-purpose variant) and three complementary benchmarks, MATH500, GPQA-diamond, and HumanEval, BoK samplers act as a practical decoding-time regulariser that improves multi-sample generation without any extra training or external verifiers. Sweeping temperature from near-deterministic ($\tau{=}0.1$) to highly stochastic ($\tau{=}0.9$), BoK samplers consistently match or outperform standard sampling and Top-K, with the largest gains precisely where vanilla sampling is most diverse but least reliable. For example, on the math-specialised model at $\tau{=}0.9$ on MATH500, BoK raises accuracy from 53.0\% (Base) to 71.6\% (+18.6\%), exceeding Top-K (56.2\%) by +15.4\%; similar high-temperature gains appear on GPQA (+6.06\%) and HumanEval (+14.64\%). These improvements are robust across a range of $(\beta,\lambda)$ choices (coverage vs. KL anchoring), indicating a stable operating region rather than brittle tuning. Finally, BoK’s benefits come with only modest compute overhead: using 5 mirror-ascent steps per token adds about 1s on MATH500 (16.88s vs. 15.84s), while even 2 steps already yield a sizeable jump (64.4\%→69.6\%) with negligible runtime increase—suggesting fast solver convergence and making BoK viable as a lightweight drop-in for practical decoding.

In short, our contributions can be stated as: 
\begin{exampleboxthree}{Summary of Contributions}
\begin{enumerate}
    \item \textbf{Unified Theory of Decoding:} We unify decoding strategies into a single framework, proving they are closed-form optima of objectives on the simplex.
    \item \textbf{Generative "Master Key" for Decoder Design:} A general framework for automatically deriving iterative algorithms for any decoding behaviour expressed as a regulariser, enabling optimisation beyond closed-form solutions.
    \item \textbf{Case Study - Best-of-K (BoK) Decoding:} We introduce Best-of-K (BoK), a coverage-based objective for self-consistency and reranking. It replaces heuristic sampling with a mathematically grounded approach that improves performance.
\end{enumerate}
\end{exampleboxthree}
\section{Decoding, Sampling and Optimisation} 

Our central claim is that decoding strategies are not heuristics, but rather implicit solutions to well-defined optimisation problems. Sampling-based methods do not introduce randomness arbitrarily; instead, they correspond to soft or regularised forms of optimisation. Conversely, deterministic methods such as greedy arise as limiting or approximate solutions to hard optimisation objectives. From this perspective, decoding algorithms differ \emph{not in kind}, but in what they optimise and the constraints they impose.

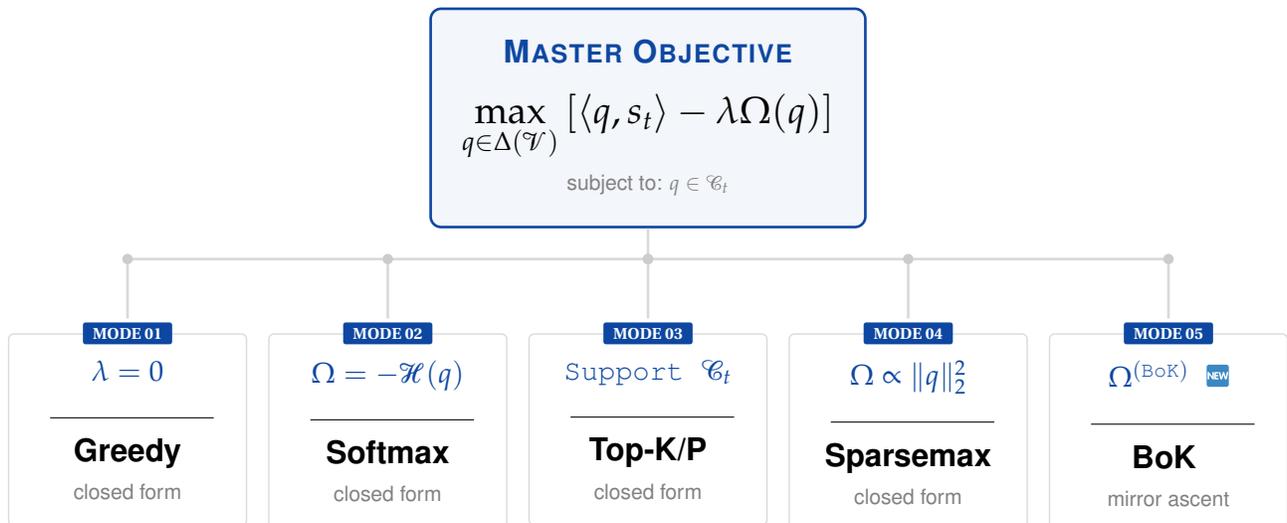
\begin{figure}[htbp]
    \centering
    \begin{adjustbox}{width=\textwidth}
    \begin{tikzpicture}[
        font=\sffamily,
        >=Stealth,
        master/.style={
            rectangle, rounded corners=4pt, draw=myblue,
            fill=myblue!5, line width=1pt, inner sep=12pt, align=center
        },
        branch/.style={
            rectangle, rounded corners=2pt, draw=gray!30,
            fill=white, inner sep=10pt, align=center, minimum width=3.2cm, line width=0.5pt,
            minimum height=1.6cm
        },
        tab/.style={
            rectangle, fill=myblue, text=white, font=\tiny\bfseries,
            inner sep=2pt, minimum width=1.2cm, rounded corners=1pt
        },
        dot/.style={circle, fill=gray!40, minimum size=4pt, inner sep=0pt},
        connector/.style={draw=gray!30, line width=1pt}
    ]

        \node[master] (M) at (0,0.5) {
            {\scshape\bfseries\color{myblue} Master Objective} \\[0.3cm]
            {\large $\displaystyle \max_{q \in \Delta(\mathcal{V})} \left[\langle q, s_t \rangle - \lambda \Omega(q)\right]$} \\[0.15cm]
            {\scriptsize\color{gray} subject to: $q \in \mathcal{C}_t$}
        };

        \coordinate (split) at (0,-1.4);
        \draw[connector] (M.south) -- (split);
        \node[dot] at (split) {};

        \draw[connector] (-7,-1.4) -- (7,-1.4);

        \node[branch, anchor=north] (G) at (-7,-2.4) {
            {\color{myblue}\small\ttfamily $\lambda = 0$} \\
            \vspace{2pt} \rule{2.2cm}{0.4pt} \\
            {\bfseries Greedy} \\
            {\scriptsize\color{gray} closed form}
        };
        \node[tab] at (G.north) {MODE 01};

        \node[branch, anchor=north] (S) at (-3.5,-2.4) {
            {\color{myblue}\small\ttfamily $\Omega = -\mathcal{H}(q)$} \\
            \vspace{2pt} \rule{2.2cm}{0.4pt} \\
            {\bfseries Softmax} \\
            {\scriptsize\color{gray} closed form}
        };
        \node[tab] at (S.north) {MODE 02};

        \node[branch, anchor=north] (T) at (0,-2.4) {
            {\color{myblue}\small\ttfamily Support $\mathcal{C}_t$} \\
            \vspace{2pt} \rule{2.2cm}{0.4pt} \\
            {\bfseries Top-K/P} \\
            {\scriptsize\color{gray} closed form}
        };
        \node[tab] at (T.north) {MODE 03};

        \node[branch, anchor=north] (P) at (3.5,-2.4) {
            {\color{myblue}\small\ttfamily $\Omega \propto \|q\|_2^2$} \\
            \vspace{2pt} \rule{2.2cm}{0.4pt} \\
            {\bfseries Sparsemax} \\
            {\scriptsize\color{gray} closed form}
        };
        \node[tab] at (P.north) {MODE 04};

        \node[branch, anchor=north] (B) at (7,-2.4) {
            {\color{myblue}\small\ttfamily $\Omega^{(\text{BoK})}$ \twemoji{new}} \\
            \vspace{2pt} \rule{2.2cm}{0.4pt} \\
            {\bfseries BoK} {\footnotesize\color{gray} } \\
            {\scriptsize\color{gray} mirror ascent}
        };
        \node[tab] at (B.north) {MODE 05};

        \foreach \pos in {-7, -3.5, 0, 3.5, 7} {
            \draw[connector] (\pos,-1.4) -- (\pos,-2.2);
            \node[dot] at (\pos,-1.4) {};
        }

    \end{tikzpicture}
    \end{adjustbox}
    \caption{Framework of Decoding as Optimisation: The master objective generalises standard LLM decoding strategies. By choosing appropriate $\lambda, \Omega(q)$ and $\mathcal{C}_t$, we can recover current decoding strategies as special cases.}
    \label{fig:master_objective}
\end{figure}

The goal of this section is to make this connection explicit. We begin by formalising decoding as the problem of selecting a distribution over next tokens, rather than directly selecting a token. This shift allows us to express decoding as an optimisation problem over distributions, balancing model score and regularisation. We then show that common decoding strategies emerge naturally as special cases. %
\subsection{Decoding as a Decision Over Distributions} 
To make the optimisation perspective precise, we start by formalising what a decoder actually does at a single generation step. Consider a language model with parameters $\bm{\theta}$. Given a prefix $\bm{x}_{<t}=(x_1, \dots, x_{t-1})$, the model produces a real-valued score for every token $v$ in the vocabulary $\mathcal{V}$. In practice, these scores are the logits or log-probabilities produced by the model. We denote them by: $s_t(v) \in \mathbb{R}$, for $v \in \mathcal{V}$. 

At this point, most decoding descriptions implicitly assume that we must immediately choose a single token $x_t$. Instead, in this work, we take a small but crucial conceptual step back. Rather than asking which token to select, we ask:
\begin{align*}
    \textcolor{blue}{{\text{Which \emph{\underline{distribution over tokens}} should we use to make the selection?}}}
\end{align*}
Formally, we introduce an auxiliary distribution $q_t(\cdot) \in \Delta(\mathcal{V})$ with $\Delta(\mathcal{V})$ denoting the probability simplex over the vocabulary $\mathcal{V}$. Once $q_t(\cdot)$ is chosen, decoding proceeds by either:
\begin{itemize}
    \item sampling $x_t \sim q_t(\cdot)$, or 
    \item deterministically selecting $x_t \in \arg\max_{v \in \mathcal{V}} q_t(v)$.
\end{itemize}
Importantly, this viewpoint is general enough to cover both stochastic and deterministic decoding, whereby \textit{i)} greedy decoding corresponds to a degenerate distribution $q_t(\cdot)$ that places all its mass on a single token \footnote{Please note we assume a rule that breaks ties.}, while \textit{ii)} sampling-based decoding corresponds to a non-degenerate $q_t(\cdot)$ with positive entropy. To illustrate this distinction, consider a simple example with a vocabulary of three tokens: $\mathcal{V} = \{a,b,c\}$. Suppose that, for a given prefix $x_t$, the language model scores each token as follows: 
\begin{align*}
    s_t(a) = 3, \quad s_t(b) = 2, \quad s_t(c) = 0. 
\end{align*}
Now, greedy decoding selects the highest-scoring token: 
\begin{equation*}
    a \in \arg\max_{v \in  \mathcal{V}} s_t(v). 
\end{equation*}

In our framework, this corresponds to choosing a degenerate distribution $q_t(\cdot)$ that places all its probability mass on the maximiser of the score function:
\begin{align*}
q_t(v) =
\begin{cases}
1, & \text{if } v \in \arg\max\limits_{u} \, s_t(u), \\
0, & \text{otherwise.}
\end{cases}
\end{align*}
In the example above, since $s_t(a) > s_t(b) > s_t(c)$, this reduces to: 
\begin{align*}
    q_t(a) = 1, \quad q_t(b) = 0, \quad q_t(c) = 0. 
\end{align*}
By contrast, sampling-based decoding selects a non-degenerate distribution that assigns nonzero probability to multiple tokens, while still favouring higher-scoring ones. In our framework, such a distribution is also obtained as a function of the score vector $s_t$. Concretely, for a given temperature parameter $\tau > 0$, sampling-based decoding chooses: 
\begin{equation*}
    q_t(v) = \frac{\exp(s_t(v)/\tau)}{\sum_{u \in \mathcal{V}}\exp(s_t(u)/\tau)}, 
\end{equation*}
which assigns higher probability to tokens with larger scores, while maintaining positive entropy. In the example above, with scores $s_{t}(a) = 3$, $s_t (b) =2$, and $s_t(c) = 0$, this construction yields a distribution of the form: 
\begin{align*}
    q_t(a) \approx 0.6, \quad \ q_t(b) \approx  0.3, \quad \ \ \ q_t(c) \approx 0, 
\end{align*}
for an appropriate choice of $\tau$. As such, sampling from $q_t (\cdot)$ will most often select the highest-scoring token $a$, but will occasionally produce $b$ or $c$. This stochasticity allows the decoder to explore alternative continuations while remaining aligned with the model’s scoring function.

Up to this point, common decoding strategies such as greedy decoding and temperature-based sampling may still appear as largely heuristic choices: practical rules that users select and tune based on empirical behaviour or trial-and-error. From this perspective, the decoder is seen as a sampler whose properties, e.g., determinism, diversity, or randomness, are controlled by externally chosen parameters such as temperature. In the next section, we show that this view is \emph{incomplete if not misleading}. In fact, these decoding strategies are not heuristic procedures, but rather exact solutions to well-defined optimisation problems. Greedy decoding and temperature-based sampling emerge naturally as optimisers of a common objective that trades off expected model score against regularisation. The apparent differences between decoding methods are therefore not algorithmic accidents, but reflect differences in the underlying objectives being optimised.

This perspective enables a shift in how decoding strategies are designed and understood. Rather than treating samplers as ad-hoc mechanisms, we can view them as solutions to optimisation problems over distributions. Under this lens, designing a decoder with desired properties such as increased diversity, robustness, or constraint satisfaction reduces to specifying the appropriate optimisation objective and constraints. We now make this connection explicit.
\subsection{Decoding as Distributional Optimisation}
The discussion so far suggests that decoding can be viewed as a decision over distributions rather than individual tokens. We now make this perspective explicit by introducing a general optimisation formulation that subsumes a wide range of decoding strategies. 

At a given decoding step $t$, recall that the model assigns a score $s_t(v)$ to each token $v \in  \mathcal{V}$. We seek to choose a distribution $q_t(\cdot) \in \Delta(\mathcal{V})$ from which the next token will be drawn (or whose mode will be selected). We define decoding as the solution to the following optimisation problem:

\begin{equation}
\label{Eq:Master}
 \text{\scriptsize\sffamily\color{myblue}\scshape ``Master Problem''} \quad \boxed{
q_t^{\ast}
\;=\;
\arg\max_{q \in \Delta(\mathcal{V})}
\Big[
\langle q, s_t \rangle
\;-\;
\lambda \, \Omega(q)
\Big]
\quad
\text{s.t.}
\quad
q \in \mathcal{C}_t.
}
\end{equation}
In the above equation, $\langle q, s_t \rangle
\;=\;
\sum_{v \in \mathcal{V}} q(v)\, s_t(v)$, as the expected model's score under the distribution $q(\cdot)$, and $\Omega(q)$ is a regularisation functional that encodes preferences such as diversity, sparsity, or stability. Moreover, $\lambda \geq 0$ controls the strength of regularisation, and $\mathcal{C}_t$ denotes constraints on the support of $q(\cdot)$. This optimisation perspective is not restricted to token-level scores: decoding criteria can also be defined over latent spaces or task-structured representations~\cite{tomov2026task}. Similar optimisation-based reinterpretations have also been developed for other core LLM mechanisms (e.g., attention)~\cite{litman2025scaled}, reinforcing the view that standard neural primitives can often be derived as solutions to explicit regularised objectives.

From a geometric perspective, each component of the formulation plays a structural role. The simplex constraints induce boundary behaviour: tokens may become active or inactive depending on whether the optimiser settles in the interior or on a face of the feasible set. Different choices of regulariser shape this geometry in different ways: some encourage interior solutions, while others allow the optimiser to concentrate mass on lower-dimensional faces. Likewise, hard support constraints restrict the feasible region itself, carving out sub-simplices before optimisation even begins. As we will see in Section \ref{Sec:Special}, classical decoding rules correspond to particular geometric choices within this framework. 

\paragraph{Interpreting Equation \eqref{Eq:Master}.} The objective above consists of two competing terms. The first term, $\langle q, s_t \rangle$, encourages the decoder to place probability mass on tokens with high model scores. If this term were optimised alone, the resulting distribution would collapse onto the highest-scoring token. The second term, $\Omega(q)$, acts as a regulariser that penalises certain properties of the distribution $q(\cdot)$. Its role is to control the shape of the decoding distribution, for example, by encouraging diversity or limiting deviation from a reference distribution. The scalar $\lambda$ determines the trade-off between strictly optimising model score and satisfying these additional preferences. Finally, the constraint set $\mathcal{C}_t$ allows the decoder to enforce hard restrictions, such as limiting the support to a subset of tokens or excluding invalid continuations. Under this view, the design of a decoding strategy reduces to:
\begin{enumerate}
    \item What notion of quality is being optimised (through $s_t(v)$); 
    \item What properties are desired in the decoding distribution (through $\Omega(q)$); and 
    \item What hard constraints must be respected (through $\mathcal{C}_t$). 
\end{enumerate}
\paragraph{Deriving {Solution Conditions} for Equation \eqref{Eq:Master}.} Deriving the closed-form solution conditions for our general optimisation problem is relatively involved. We present it in steps that make various assumptions about $\mathcal{C}_t$, $\Delta(q)$ and $\Omega (q)$. In the first case, we consider the absence of $\mathcal{C}_t$. Here, our optimisation problem becomes the following:  
\begin{equation}
\label{Eq:SimpleOne}
    q_t^{\star} = \arg\max_{q \in \Delta(\mathcal{V})} \left(\langle q,s_t\rangle - \lambda \Omega(q)\right), \ \ \text{for $\lambda \geq 0$}, 
\end{equation}
where $\Delta(\mathcal{V})$ is defined as: $
    \Delta(\mathcal{V})=\left\{q \in \mathbb{R}^{|\mathcal{V}|}: q(v) \geq 0, \sum_{v} q(v) = 1\right\}$. It is easy to see that maximising the objective in Equation \eqref{Eq:SimpleOne} can be equivalently rewritten as a minimisation problem, such that: 
\begin{equation}
\label{Eq:ZeMin}
q_t^{\star} = \arg\min_{q \in \Delta(\mathcal{V})} (\lambda \Omega(q)- \langle q,s_t\rangle).    
\end{equation}
Assuming that $\Omega(q)$ is convex\footnote{The convexity of $\Omega(q)$ is typical. As we show later, many special cases of LLM decoding strategies stem from different convex choices of $\Omega(q)$.}, the minimisation problem we just derived (Equation \eqref{Eq:ZeMin}) is simply a ``convex - linear'' minimisation objective for which clean optimality conditions are easily attainable. What makes solving the problem in Equation \eqref{Eq:ZeMin} hard is the existence of the constraint set $\Delta(\mathcal{V})$. If we were to ignore those constraints, we could just set the derivative to zero, like in unconstrained calculus, and proceed. However, the existence of those constraints means we must be more careful: we are minimising over a restricted set, so we should take those into account. 
\begin{equation*}
    \textcolor{blue}{\text{But what do those constraints \underline{\emph{actually mean}?}}}
\end{equation*}
Our constraints in $\Delta({\mathcal{V}})$ are not fancy at all. They are pretty simple! All they do is make $q(\cdot)$ a valid probability distribution. In detail, when we say $q \in \mathcal{V}$, we really mean the following two properties should hold: 
\begin{itemize}
    \item $\sum_{v} q(v) = 1 \rightarrow$ Forcing $q(\cdot)$ to be a probability distribution that sums to 1 across all realisations of the random variable; and 
    \item $q(v) \geq 0$ for all $v \in \mathcal{V} \rightarrow$ You can't have negative $q(\cdot)$ values on any $v \in \mathcal{V}$.  
\end{itemize}
As is tradition in mathematical derivations, we begin by pretending the inconvenient part does not exist, ignore the second condition ($q(v) \geq 0$), and come back to it later. So pretend for a moment that $q(v)$ can be any real numbers as long as they sum to 1. Here, our optimisation problem becomes: 
\begin{equation*}
    \min_{q} \mathcal{J}(q) \equiv \lambda \Omega(q) - \sum_{v} q(v) s_t (v), \quad \text{s.t.}  \sum_{v} q(v) =1.
\end{equation*}
Now, we really, really want to take derivatives and set them to zero. Sadly, there is a constraint! A standard trick is: \emph{add the constraint into the objective with a ``price'' $\eta$}. This gives us what every constrained optimisation problem eventually produces, the Lagrangian:
\begin{equation*}
    \mathcal{L}(q,\eta) = {\lambda \Omega(q)}^{\textcolor{blue}{(1)}} -\sum_v q(v)s_t(v)^{\textcolor{blue}{(2)}} + \underbrace{\eta \left(\sum_v q(v) -1\right)}_{\text{The Constraint Penalty}}^{\textcolor{blue}{(3)}}. 
\end{equation*}
We now have an unconstrained objective, which allows us to deploy one of the great discoveries of the last century: the gradient. Let us do that! Let us take the partial derivatives of $\mathcal{L}(q,\eta)$ with respect to $q(v)$. Doing so, term-by-term, we get: 
\begin{align*}
   \textcolor{blue}{ (1) \rightarrow} \  \frac{\partial}{\partial q(v)}\lambda \Omega(q) = \lambda \frac{\partial}{\partial q(v)} \Omega(q)(v)&, \quad \textcolor{blue}{(2) \rightarrow} \ \frac{\partial}{\partial q(v)}\left(-\sum_v q(v)s_t(v)\right) = -s_t(v), \\ 
    &\hspace{-5em} \textcolor{blue}{(3) \rightarrow} \ \frac{\partial}{\partial q(v)} \eta (\sum_v q(v) -1) = \eta.
\end{align*}

So on the stationary point $q_t^{\star}$, we get this condition: 
\begin{align}
\label{Eq:Condition}
    \lambda \frac{\partial}{\partial q(v)} \Omega_{q(v)}(q^{\star}_t)(v)  - s_t(v) + \eta = 0 \quad \implies \quad s_t(v) - \lambda \frac{\partial}{\partial q(v)}\Omega(q_t^{\star})(v) = \eta,  \ \ \ \forall v \in \mathcal{V}. 
\end{align}
It is easy``ish'' to interpret the condition in Equation \eqref{Eq:Condition}. It simply says that for every token $v$, the quantity $s_t(v) - \lambda\frac{\partial}{\partial q(v)} \Omega(q_t^{\star})(v)$ must be the same constant $\eta$. This is a ``balancing'' condition created by the fact that probabilities must sum to 1. 

\paragraph{Getting the Inequality Back.} At this stage, we have derived a general condition, but we should not forget about our previous promise! We still have to enforce $q(v) \geq 0$ because some solutions to Equation \eqref{Eq:Condition} can give negative $q_t^{\star}(v)$. 
\begin{exampleboxtwo}{One Dimensional Example for Handling Inequality Constraints}
For now, let us forget about tokens and consider a single-variable objective:
\begin{equation*}
    \min_{x \geq 0} f(x).
\end{equation*}
An optimal $x^{\star}$ means that \emph{no feasible small move} from $x^{\star}$ can reduce the function value. 
A move is \emph{feasible} if it respects the constraint $x\ge 0$. Thus:
\textit{i)} if $x^{\star}>0$ (an interior point), we can move a small amount both to the right and to the left, i.e., to $x^{\star}+\delta$ and $x^{\star}-\delta$, for sufficiently small $\delta>0$;
\textit{ii)} if $x^{\star}=0$ (a boundary point), we can only move to the right (to $0+\delta$), since $0-\delta$ would violate $x\ge 0$. To make this precise, we use a first-order Taylor approximation. For small $\delta>0$,
\begin{align*}
    f(x^{\star}+\delta) &\approx f(x^{\star}) + \delta f'(x^{\star}), \\
    f(x^{\star}-\delta) &\approx f(x^{\star}) - \delta f'(x^{\star}).
\end{align*}
Since $x^{\star}$ is optimal, any feasible move must not decrease $f$, i.e.,
\begin{equation*}
    f(x^{\star}+\delta)-f(x^{\star}) \ge 0 \quad \text{for all sufficiently small feasible }\delta>0,
\end{equation*}
and, when $x^{\star}>0$ (so that $x^{\star}-\delta$ is also feasible),
\begin{equation*}
    f(x^{\star}-\delta)-f(x^{\star}) \ge 0 \quad \text{for all sufficiently small }\delta>0.
\end{equation*}
Combining these inequalities with the Taylor expansions gives the two cases below:
\begin{itemize}
    \item \textbf{Case I (Interior optimum $x^{\star}>0$):} both $x^{\star}+\delta$ and $x^{\star}-\delta$ are feasible. Hence
    \[
        \delta f'(x^{\star}) \ge 0 \quad \text{and} \quad -\delta f'(x^{\star}) \ge 0 \quad \forall \delta>0,
    \]
    which implies $f'(x^{\star})=0$.
    \item \textbf{Case II (Boundary optimum $x^{\star}=0$):} only $0+\delta$ is feasible. Hence
    \[
        \delta f'(0) \ge 0 \quad \forall \delta>0,
    \]
    which implies $f'(0)\ge 0$.
\end{itemize}
In other words, these conditions reflect whether $x^{\star}$ lies in the interior or on the boundary of the feasible set. 
If $x^{\star}>0$, we can perturb $x^{\star}$ both to the right and to the left while remaining feasible, and optimality therefore forces the slope to vanish, i.e., $f'(x^{\star})=0$. 
If $x^{\star}=0$, we can only move ``into'' the feasible region (to the right), so optimality only requires that rightward perturbations do not decrease $f(\cdot)$, equivalently $f'(0)\ge 0$ (a one-sided slope condition).
\end{exampleboxtwo}
Having understood what conditions we would need when having a non-negativity feasibility set, we can now go back to our tokens and apply the same reasoning coordinate-wise to each component $q(v)\ge 0$. In our problem, the role of $f'(x^{\star})$ is played by the partial derivative of the Lagrangian with respect to the coordinate $q(v)$. Recall the Lagrangian:
\[
\mathcal{L}(q,\eta) \;=\; \lambda \Omega(q) - \langle q,s_t\rangle + \eta\Big(\sum_{u\in \mathcal{V}} q(u)-1\Big),
\]
where the scalar $\eta$ enforces the normalisation constraint $\sum_{u} q(u)=1$. Taking the partial derivative with respect to $q(v)$ yields:
\[
\frac{\partial }{\partial q(v)}\mathcal{L}(q,\eta)
\;=\;
\lambda\frac{\partial }{\partial q(v)}\Omega(q)(v) - s_t(v) + \eta.
\]
By the same one-dimensional argument as above, optimality under the constraint $q(v)\ge 0$ implies a \emph{two-case} condition:
\[
\begin{cases}
q^{\star}(v)>0 \;\;\Rightarrow\;\; \frac{\partial}{\partial q(v)}\mathcal{L}(q^{\star},\eta)=0,\\[4pt]
q^{\star}(v)=0 \;\;\Rightarrow\;\; \frac{\partial }{\partial q(v)}\mathcal{L}(q^{\star},\eta)\ge 0.
\end{cases}
\]
Substituting the expression for $\frac{\partial}{\partial q(v)} \mathcal{L}(\cdot, \cdot)$ and rearranging gives us two conditions for the optimality of the solution, which we state below. 

\begin{exampleboxthree}{Take Home Message: KKT Optimality Conditions}
From the above derivations, the take-home message is that we can characterise an optimal $q_t^{\star}$ using the following identities:  
    \begin{equation}
    \label{Eq:Conds}
    \begin{alignedat}{2}
        &\text{\scriptsize\sffamily\color{myred}\scshape GLOBAL} \quad && \sum_{v \in \mathcal{V}} q_t^\star(v) = 1, \quad q_t^\star(v) \ge 0 \quad \forall v \in \mathcal{V}. \\[1pt]
        &\text{\scriptsize\sffamily\color{myblue}\scshape ACTIVE} \quad && q_t^\star(v) > 0 \ \implies \ s_t(v) - \lambda \frac{\partial}{\partial q(v)} \Omega(q_t^{\star})(v) = \eta, \\[1pt]
        &\text{\scriptsize\sffamily\color{gray}\scshape INACTIVE} \quad && q_t^\star(v) = 0 \ \implies \ s_t(v) - \lambda \frac{\partial}{\partial q(v)} \Omega(q_t^{\star})(v) \le \eta.
    \end{alignedat} 
   \end{equation}
\end{exampleboxthree}
\section{LLM Decoding Strategies are Different Regularisers} \label{Sec:Special}
At this point, the reader may reasonably wonder what all of the preceding mathematics was for. We now reap what we have sown (Galatians 6:7). In this section, we show that many widely used decoding strategies arise as simple special cases of our general formulation. Those will correspond to different choices of regularisation $\Omega(q)$, $\lambda$ and $\mathcal{C}_t$. 
\subsection{Greedy Decoding: The Boring but Necessary Case}
We begin with the simplest possible case, which serves mainly to reassure us that nothing has gone terribly wrong. To recover greedy decoding, we assume there is no regulariser by setting $\lambda = 0$. We also ignore $\mathcal{C}_t$ for the time being. As such, our ``Master'' objective from Equation \eqref{Eq:Master} quietly collapses into a far more modest one: 
\begin{equation*}
    q_t^{\star} = \arg\max_{q \in \Delta(\mathcal{V})} \langle q, s_t\rangle \equiv \arg\max_{q \in \Delta(\mathcal{V})} \sum_{v } q(v) s_t(v). 
\end{equation*}

We now “cash in” the take-home KKT-style conditions from Equation \eqref{Eq:Conds}. In this greedy case, we have for some scalar $\eta$: 
\[
\text{\scriptsize\sffamily\color{myblue}\scshape ACTIVE}\ \ q_t^\star(v)>0 \ \Rightarrow\ s_t(v)=\eta,
\qquad
\text{\scriptsize\sffamily\color{gray}\scshape INACTIVE}\ \ q_t^\star(v)=0 \ \Rightarrow\ s_t(v)\le \eta.
\]
These two lines already characterise the solution: every token that receives a nonzero probability
must have \emph{exactly the same} score, and every token with zero probability must have a score no larger. Let $M := \max_{u\in \mathcal{V}} s_t(u)$ and define the argmax set
\[
\mathcal{V}^\star := \arg\max_{u\in \mathcal{V}} s_t(u)=\{v\in \mathcal{V}: s_t(v)=M\}.
\]
The inactive condition forces $\eta \ge s_t(v)$ for all $v$, hence $\eta \ge M$.
But the active condition says there exists at least one active token (since $\sum_v q_t^\star(v)=1$),
and any active token must satisfy $s_t(v)=\eta$, hence $\eta$ must equal the maximum score: $\eta = M.$ Therefore, the only tokens that can be active are those in $\mathcal{V}^\star$, i.e.,
\[
q_t^\star(v)>0 \ \Rightarrow\ v\in \mathcal{V}^\star,
\qquad\text{equivalently}\qquad
\operatorname{supp}(q_t^\star)\subseteq \mathcal{V}^\star.
\]
In words: \emph{an optimal solution places all probability mass on the highest-scoring token(s).}
If the maximiser is unique, $\mathcal{V}^\star=\{v^\star\}$, then the unique optimum is the degenerate distribution
\[
q_t^\star(v)=\delta_{v^\star}(v),
\qquad\text{where}\qquad
v^\star\in\arg\max_{u\in \mathcal{V}} s_t(u).
\]
If there are ties (i.e., $|\mathcal{V}^\star|>1$), then \emph{any} distribution supported on $\mathcal{V}^\star$ is optimal. In other words,
greedy decoding corresponds to selecting a vertex of this optimal face, i.e., picking one $v^\star\in \mathcal{V}^\star$
and using $q_t^\star=\delta_{v^\star}$.

\subsection{From Negative Entropy to Softmax: A Predictable Ending} \label{Sec:Entropy}
Let us now consider the case where, in a move that will surprise no one, $\Omega(q)$ is chosen to be the negative entropy of $q$, such that: $\Omega(q) = \sum_{v \in \mathcal{V}} q(v) \log q(v).$ This will make our optimisation objective from Equation \eqref{Eq:Master} look like the following: 
\begin{equation*}
    q_t^{\star} = \arg\max_{q \in \Delta(\mathcal{V})}\left[\sum_v q(v)s_t(v) - \lambda \sum_v q(v) \log q(v)\right]. 
\end{equation*}
Returning to the conditions in Equation \eqref{Eq:Conds}, the suspense is, of course, which case applies: active or inactive? This hinges on a single question: does $q^{\star}_t(\cdot)$ live comfortably in the interior of the simplex, or does it press up against the boundary?
\begin{exampleboxtwo}{The Key Mathematical Fact: The Derivative Blows Up at Zero}
To understand this, let us take a step back and look at the derivative of $f(x) = x \log x$. Why this $f(x)$? That is how the entropy looks! For $x > 0$, we can see that: 
\begin{align*}
    \frac{d}{dx}(x\log x) &= \frac{d}{dx}x\times\log x + \frac{d}{dx}\log x \times x \\
    & = 1 \times \log x + \frac{1}{x} \times x \\
    & = 1 + \log x.
\end{align*}
We now examine what happens as $x$ approaches the mysterious value $0^{+}$: 
\begin{equation*}
    \lim_{x\rightarrow 0^{+}} 1 + \log x \rightarrow -\infty,
\end{equation*}
because as $x \rightarrow 0^+$, $\log x \rightarrow - \infty$, and thus $1+\log x \rightarrow - \infty$. In other words, as our variable approaches $0^{+}$, the gradients blowup. Therefore, if we were optimising for $x$, every time we get closer to zero, the gradient politely but firmly tells us to move.
\end{exampleboxtwo}
Equipped with this realisation, we return to our case that contains the negative entropy regulariser $\Omega(q)$, which strongly discourages zero probabilities. Conveniently, the gradients make this preference explicit: 
\begin{equation}
\label{Eq:Der}
    \frac{\partial }{\partial q(v)} \Omega(q) = 1 + \log q(v),
\end{equation}
which blows up as we $q(v)$ approaches zero. Hence, our winner is \twemoji{drum} ... the \text{\scriptsize\sffamily\color{myblue}\scshape ACTIVE} case: 
\begin{equation*}
   \text{\scriptsize\sffamily\color{myblue}\scshape ACTIVE} \quad q_t^\star(v) > 0 \ \implies \ s_t(v) - \lambda \frac{\partial}{\partial q(v)} \Omega(q_t^{\star})(v) = \eta. 
\end{equation*}
Let us do some algebra! Replacing the derivative from Equation \ref{Eq:Der} in our condition, we get: 
\begin{align*}
    s_t(v) - \lambda (1 + \log q_t^{\star}(v)) - \eta =0  &\implies \lambda (1 + \log q_t^{\star}(v)) = s_t(v) - \eta \\
    & \implies \log q^{\star}_t(v) = \frac{s_t(v) - \eta}{\lambda} - 1 \\ 
    & \implies q^{\star}_t(v) = \exp\left(\frac{s_t(v) - \eta}{\lambda} - 1 \right) \\ 
   & = C \exp{\left(\frac{s_t(v)}{\lambda}\right)},
\end{align*}
with the constant $C = e^{-1} e^{-\frac{\eta}{\lambda}}$. We know that $q_t^\star(v)$ must be a valid probability distribution, and valid probability distributions insist on summing to one. Let us indulge them: 
\begin{align*}
    \sum_{u \in \mathcal{V}} q_t^{\star}(u) = 1 \implies  \sum_{u \in \mathcal{V}}  C \exp{\left(\frac{s_t(u)}{\lambda}\right)} = 1 &\implies C \sum_{u \in \mathcal{V}} \exp{\left(\frac{s_t(u)}{\lambda}\right)} =1 \\ 
    & \implies  C = \frac{1}{\sum_{u \in \mathcal{V}} \exp{\left(\frac{s_t(u)}{\lambda}\right)}}.
\end{align*}
Therefore, our optimal distribution $q^{\star}_t(\cdot)$ if we picked $\Omega(q)$ to be the negative entropy regulariser, ends up being: 
\begin{equation*}
 \text{\scriptsize\sffamily\color[RGB]{128,0,32}\scshape SoftMax Decoders: } \quad   q_t^{\star}(v) = \frac{\exp{\left({s_t(v)}/{\lambda}\right)}}{\sum_{u \in \mathcal{V}} \exp{\left({s_t(u)}/{\lambda}\right)}}. 
\end{equation*}
We thus recover the classical softmax decoder directly from our master optimisation problem. In particular, choosing the (negative) Shannon entropy as the regulariser produces the familiar temperature-controlled distribution, with $\lambda$ playing exactly the role of the temperature 
$\tau$ in the LLM literature.

This demonstrates that softmax decoding is not an independent heuristic, but the optimiser of our abstract objective under an entropic geometry. In this sense, our framework elevates decoding to a higher-level problem definition: different decoders arise from different regularisers. Decoder design becomes regulariser design.
\subsection{Trimming the Vocabulary: Top-K Samplers} 
Fortunately, extending the results derived above to Top-K samplers requires only minimal additional work. Until now, we have been pretending that the constraint $\mathcal{C}_t$ in Equation~\ref{Eq:Master} does not exist. For Top-K, this pretence is no longer sustainable. First, we define the indices of the $k$ highest-scoring tokens. We define the set $\mathcal{V}_k \subset \mathcal{V}$ such that: \textit{i)} $|\mathcal{V}_k| = k$, and \textit{ii)} $\forall v \in \mathcal{V}_k$ and $\forall u \notin \mathcal{V}_k$, $s_t(v) \ge s_t(u)$. This allows the constraint set $\mathcal{C}_t^{(k)}$ to be a subset of the simplex $\Delta(\mathcal{V})$ where tokens outside the top-$k$ are forced to have zero probability:
\begin{equation*}
\mathcal{C}_t^{(k)} = \{ q \in \Delta(\mathcal{V}) : q(v) = 0, \forall v \notin \mathcal{V}_k \}.     
\end{equation*}

To ensure we end up with a Softmax over the Top-K candidates, as before, we use the (negative) Shannon entropy as our regulariser $\Omega(q) = \sum_{v \in \mathcal{V}} q(v) \log q(v)$. With this, the objective becomes:
\begin{equation*}
q_{t}^{*} = \arg \max_{q \in \mathcal{C}_t^{(k)}} \left[ \sum_{v \in \mathcal{V}} q(v) s_t(v) - \lambda \sum_{v \in \mathcal{V}} q(v) \log q(v) \right].    
\end{equation*}
Because $q(v) = 0$ for all $v \notin \mathcal{V}_k$ per our constraint, the summation effectively collapses to the indices in $\mathcal{V}_k$: 
\begin{equation*}
q_{t}^{*} = \arg \max_{q \in \Delta(\mathcal{V}_k)} \left[ \sum_{v \in \mathcal{V}_k} q(v) s_t(v) - \lambda \sum_{v \in \mathcal{V}_k} q(v) \log q(v) \right].    
\end{equation*}
Following a similar derivation to that in Section \ref{Sec:Entropy}, we arrive at Top-K samplers giving us: 
\begin{equation*}
\text{\scriptsize\sffamily\color[RGB]{128,0,32}\scshape Top-K Decoders: } \quad   q_t^{\star}(v) = 
\begin{cases} 
\frac{\exp(s_t(v)/\lambda)}{\sum_{u \in \mathcal{V}_k} \exp(s_t(u)/\lambda)} & \text{if } v \in \mathcal{V}_k, \\ 
0 & \text{otherwise. } 
\end{cases}  
\end{equation*}

\subsection{Mind the Mass: Top-P Sampling}
Top-P (nucleus) sampling adapts the size of the active support set based on the model's
confidence in the current context. When the model's distribution is \emph{flat} (high uncertainty),
the nucleus expands to preserve diversity; when it is \emph{peaky} (high confidence), the nucleus
contracts, filtering out low-probability noise in the long tail.

In our optimisation view, Top-P is obtained by replacing the fixed-cardinality support constraint
of Top-K with a \emph{cumulative-mass support constraint}. Rather than restricting the decoder to
exactly K tokens irrespective of confidence, Top-P restricts the decoder to the smallest set of
tokens whose \emph{model-assigned probability mass} exceeds a threshold $p\in(0,1]$.

\paragraph{Defining the nucleus.}
Let $p_t(\cdot)$ denote the model distribution over next tokens induced by the scores $s_t$. Sort tokens in descending order of $p_t$: $
p_t(v^{(1)}) \ge p_t(v^{(2)}) \ge \cdots.$ Define $m$ to be the smallest index such that the cumulative mass reaches $p$, i.e., $\sum_{i=1}^{m} p_t\!\left(v^{(i)}\right) \ge p$, and set the nucleus to be\footnote{If ties occur at the cutoff, we may break them arbitrarily.}:
\[
\mathcal V_p := \{v^{(1)},\dots,v^{(m)}\}.
\]

This induces the context-dependent constraint set $
\mathcal C_t^{(p)} := \{q\in\Delta(\mathcal V): q(v)=0,\ \forall v\notin \mathcal V_p\}$, which is a sub-simplex of $\Delta(\mathcal V)$ supported on $\mathcal V_p$. To recover the standard Top-P sampler, we again use the (negative) Shannon entropy regulariser, and solve the master objective restricted to $\mathcal C_t^{(p)}$:
\[
q_t^\star = \arg\max_{q\in \mathcal C_t^{(p)}} \Big[ \langle q, s_t\rangle - \lambda \Omega(q)\Big].
\]                           
Because $q(v)=0$ for all $v\notin \mathcal V_p$, this optimisation is equivalent to optimising over
the simplex $\Delta(\mathcal V_p)$:
\[
q_t^\star = \arg\max_{q\in \Delta(\mathcal V_p)} \Big[ \sum_{v\in\mathcal V_p} q(v)s_t(v)
- \lambda \sum_{v\in\mathcal V_p} q(v)\log q(v)\Big].
\]
The derivation in Section~\ref{Sec:Entropy} therefore applies verbatim, yielding a softmax distribution renormalised over the nucleus:
\[
\text{\scriptsize\sffamily\color[RGB]{128,0,32}\scshape Top-P Decoders: } \quad q_t^\star(v)=
\begin{cases}
\dfrac{\exp\!\left(s_t(v)/\lambda\right)}{\sum\limits_{u\in \mathcal V_p}\exp\!\left(s_t(u)/\lambda\right)}
& \text{if } v\in \mathcal V_p,\\[10pt]
0 & \text{otherwise.}
\end{cases}
\]
In words, Top-P sampling first selects a context-dependent nucleus $\mathcal V_p$ based on
cumulative model probability mass, and then samples from a temperature-controlled softmax
restricted to that nucleus.

\subsection{Letting Probabilities Go to Zero: Sparsemax Decoding}
While standard Softmax sampling is the standard choice for text generation, it suffers from the ``heavy tail'' problem: because the derivative of Shannon entropy approaches $-\infty$ as any probability $q(v)$ approaches zero, the optimiser is strictly forbidden from reaching the boundary of the simplex. This forces the model to assign a non-zero, albeit small, probability to every single token in the vocabulary, which can lead to the ``tail risk'' of sampling nonsensical or hallucinatory tokens. 

To solve this without the ad-hoc truncation rules of Top-K or Top-P, we look toward Sparsemax \cite{martins2016softmax}. By replacing the logarithmic penalty of entropy with a quadratic penalty, we allow the optimisation to reach the simplex boundary, effectively performing an automated, adaptive truncation that assigns exactly zero probability to low-scoring candidates. We again consider a single decoding step with an empty $\mathcal{C}_t$ and solve the ``Master problem'' again. In this special case, we choose the quadratic regulariser: 
\begin{equation}
\label{Eq:Quad}
\Omega(q)=\frac12\lVert q\rVert_2^2=\frac12\sum_{v\in V} q(v)^2,
\qquad\Rightarrow\qquad q_{t}^{\star} = \arg\max_{q \in \Delta(\mathcal{V})}\left[\langle q,s_t \rangle - \frac{\lambda}{2} ||q||_2^{2}\right].
\end{equation}
Please note that in our problem, both conditions in Equation \eqref{Eq:Conds} need to be considered. This is the case since $\Omega(q)= \frac{1}{2}||q||_{2}^{2}$ allows solutions that are positive but can also equate to zero. Of course, the latter condition is what enables sparsity. Let us first consider the active tokens case, i.e., when $q^{\star}_t (v) > 0$. For a token $v \in \mathcal{V}$, the active condition can be written as: 
\begin{align*}
   \text{\scriptsize\sffamily\color{myblue}\scshape ACTIVE} \quad   s_t(v) - \lambda \frac{\partial}{\partial q(v)}\Omega(q^{\star}_t)(v)  = \eta  \quad &\implies \quad 
   s_t(v) - \lambda \frac{\partial}{\partial q(v)}\left[\frac{1}{2}\sum_{v\in\mathcal{V}} q_t^{\star}(v)^2\right]   = \eta \\
   &\implies s_t(v) - \lambda q_t^{\star}(v) = \eta.
\end{align*}
Solving for $q^{\star}(v)$, we get: $q^{\star}_t(v) = \frac{1}{\lambda}(s_t(v) - \eta)$. Also notice that since we are in the $q_{t}^{\star}(v)>0$, we can, thus, clearly see that $s_t(v) > \eta$. 

Moving on to the second condition, we observe the following: 
\begin{equation*}
  \text{\scriptsize\sffamily\color{gray}\scshape INACTIVE} \quad   s_t(v) - \lambda \frac{\partial}{\partial q(v)} \Omega(q_t^{\star})(v) \leq \eta \quad \implies \quad s_t(v) - \lambda q_t^{\star}(v) \leq \eta . 
\end{equation*}
Since this condition holds for a token in which $q^{\star}_t(v) = 0$, then we can conclude that: $s_t(v) \leq \eta$. So, essentially, from those two conditions we learned that: 
\begin{equation*}
q_t^\star(v)=
\begin{cases}
\dfrac{s_t(v)-\eta}{\lambda} & \text{if } s_t(v)>\eta \\[6pt]
0 & \text{if } s_t(v)\le \eta.
\end{cases}    
\quad \implies \quad q_{t}^{\star}(v) = \frac{1}{\lambda} [s_t(v) - \eta]_{+},  
\end{equation*}
\text{with}$ \quad [x]_+ = \max(0,x) = \text{ReLU}(x)$. The last remaining ingredient needed to finalise our derivation is determining $\eta$. To do so, we make use of the constraint that $\sum_{v \in \mathcal{V}} q_t^{\star}(v)  = 1$: 
\begin{align*}
    \sum_{v \in \mathcal{V}} q^{\star}_t(v) = 1 \quad \implies \quad  \sum_{v \in \mathcal{V}} \frac{1}{\lambda} [s_t(v) - \eta]_{+} = 1 \quad &\implies \quad \sum_{v \in \mathcal{V}}[s_{t}(v) - \eta]_{+} =\lambda \\
    &\implies \quad \sum_{v\in \mathcal{V}} \max(0, s_t-\eta) = \lambda. 
\end{align*}
In other words, we need $\eta$ such that the sum of the positive parts equals $\lambda$. As we see, it is challenging to acquire a closed-form solution for $\eta$. Therefore, we develop an algorithmic solution. The key idea is that we will assume we know which tokens are active by defining an active set $\mathcal{S}(\eta)=\{v \in \mathcal{V}: s_t(v) > \eta\}$. Then, we have: \textit{i)} If $v \in \mathcal{S}(\eta)$, then $[s_t(v)-\eta]_{+} = s_t(v) - \eta$; and \textit{ii)} If $v \notin \mathcal{S}(\eta)$, then $[s_t(v)-\eta]_{+} = 0$. Therefore, those conditions over $\eta$ simplify to a sum over the active set $\mathcal{S}(\eta)$, giving us: $\sum_{v \in \mathcal{V}}(s_{t}(v)-\eta) = \lambda$. Unfortunately, this result is still ``implicit'' because $\mathcal{S}(\eta)$ depends on $\eta$ itself. So, we apply the standard trick of guessing the active set size $k$. 

The key observation is that the
\emph{active} tokens are exactly those satisfying $s_t(v)>\eta$. Consequently, if a token with some score $s_t(v)$ is active, then every token with a larger score must also be active. Therefore, the support of $q^\star_t(\cdot)$ must be a \emph{Top-K} set for some $k$.

Let $s_t^{(1)}\ge s^{(2)}_t\ge \cdots \ge s^{(n)}_t$ denote the scores sorted in descending order, where $n:=|\mathcal{V}|$.
Assume that the active set has size $k$ and is given by the top-$k$ indices:
\[
\mathcal{S}_k(\eta) = \{(1),(2),\ldots,(k)\}.
\]
Under this hypothesis, the normalisation condition becomes: $$
\sum_{v\in \mathcal{S}(\eta)} q_t^\star(v)=\frac{1}{\lambda}\sum_{i=1}^k \big(s^{(i)}_t-\eta\big) = 1,$$ since tokens outside $\mathcal{S}_k(\eta)$ contribute zero after the $[\cdot]_+$ clipping. Defining the prefix sum: $
A_k := \sum_{i=1}^k s^{(i)}_t$, we obtain: 
\[
1=\frac{1}{\lambda}\big(A_k-k\eta\big)
\qquad \implies \qquad
A_k-k\eta=\lambda
\qquad \implies \qquad
\eta_k:=\frac{A_k-\lambda}{k}.
\]
Importantly, $\eta_k$ is not an arbitrary choice: it is the \emph{unique} threshold value that would make
$\sum_v q^\star_t(v)=1$ \emph{if} the active set were exactly the top $k$ tokens. The remaining question is which $k$ is self-consistent. The Top-K hypothesis holds precisely when the computed threshold
$\eta_k$ separates the top $k$ scores from the rest, i.e., $
s^{(k)}_t > \eta_k
\quad\text{and}\quad
s^{(k+1)}_t \le \eta_k$, with the convention $s^{(n+1)}_t=-\infty$. We select any $k^\star$ satisfying the above inequalities, set
$\eta=\eta_{k^\star}$, and then recover the optimal distribution via: 
\begin{equation*}    
\text{\scriptsize\sffamily\color[RGB]{128,0,32}\scshape SparseMax-Style Decoders: } \quad q_t^\star(v)=\frac{1}{\lambda}\,[s_t(v)-\eta]_+.
\end{equation*}
\section{Going Beyond Current Decoders}
So far, our story has been pleasantly ``closed-form''. Once we write decoding as a regularised optimisation problem on the simplex, the KKT conditions act like a master key, and classical decoders fall out as special cases. In a complementary direction, prior work has shown that search-based decoding methods (e.g., beam search) can also be reinterpreted as optimising an explicit regularised objective~\cite{meister2020if}, and recent work on controlled and alignment-oriented decoding also constructs inference-time policies by optimising explicit regularised objectives (often KL-anchored to the base model) rather than relying on heuristic sampling rules~\cite{mudgal2023controlled, chakraborty2024transfer}. But this view raises an immediate practical question: what do we do when the optimiser is no longer solvable in one line? The moment we introduce richer regularisers, coupling terms, coverage objectives, or more structured constraints, the distribution $q_t^{\star}(\cdot)$ is still well-defined—but we cannot always write it down analytically. This section switches from deriving decoders to \emph{computing them}. We introduce mirror descent (and mirror ascent) as a principled method tailored to simplex geometry. It preserves non-negativity and normalisation by construction and provides an algorithmic template that solves our master problem even when closed-form solutions are unavailable.
\subsection{Why not just run Vanilla (projected) Gradient Ascent on $q(\cdot)$?}
If we look back at our master problem, it is natural to try solving it with a standard tool from constrained optimisation: projected gradient ascent \cite{bertsekas1999nonlinear,boyd}. After all, we are maximising a differentiable objective over a convex set (the simplex, possibly intersected with additional constraints). One could take a gradient step 
 and then project back onto $\Delta(\mathcal{V})$, repeating until convergence. In practice, this approach is indeed feasible, and it is often the first method people reach for. To make this explicit, we write the following optimisation problem and update rule\footnote{Please notice that we have ignored $\mathcal{C}_t$ for now. We can easily incorporate it again by defining a sub-simplex depending on the $\Delta(\mathcal{V}) \subseteq\mathcal{V} \cap \mathcal{C}_t$.}: 
 \begin{align*}
    \text{\scriptsize\sffamily\color{myblue}\scshape ``Master Problem''} \quad   &\max_{q \in \Delta(\mathcal{V})} f(q), \quad \text{with} \quad f(q) = \langle q, s_t \rangle
\;-\;
\lambda \, \Omega(q), \\
\text{\scriptsize\sffamily\color{black}\scshape ``Projected Gradients'' \quad}    &\quad \quad \quad \boxed{q_{j+1} = \prod_{\Delta(\mathcal{V})} \left(q_j + \eta \nabla f(q_j)\right),} 
 \end{align*}
 with $\prod_{\Delta(\mathcal{V})}$ denoting the projection onto the simplex.

 \paragraph{What do projected gradients solve?} We note that the projected ascent step above is also somewhat misguided as it implicitly equips the simplex with an $L_2$ geometry that does not match the way probability distributions behave, especially near the boundary where many tokens should receive exactly zero (or near-zero) mass. This mismatch can lead to unstable updates, overly aggressive redistribution of mass, and “fighting the constraints” at every iteration. This is not a philosophy! We can explicitly characterise this property by understanding that projected gradients are the solution to the following optimisation problem: 
 \begin{equation}
 \label{Eq:PGASurrogate}
     q_{j+1} = \arg\max_{q \in \Delta(\mathcal{V})}\left[\langle \nabla f(q_j), q-q_j \rangle - \frac{1}{2\eta}\left|\left|q - q_j\right|\right|_{2}^{2}\right].
 \end{equation}
 Let us see why! We begin by expanding the terms in Equation \eqref{Eq:PGASurrogate}:
 \begin{align*}
      q_{j+1} &= \arg\max_{q \in \Delta(\mathcal{V})}\left[\nabla f(q_j)^{\mathsf{T}}(q - q_j)-\frac{1}{2\eta}\left(||q||_{2}^{2} - 2 \langle q, q_j \rangle + ||q_j||_2^2\right)\right] \\
      & = \arg\max_{q \in \Delta(\mathcal{V})}\left[\nabla f(q_j)^{\mathsf{T}}q - \nabla f(q_j)^{\mathsf{T}} q_j-\frac{1}{2\eta}\left(||q||_{2}^{2} - 2 \langle q, q_j \rangle + ||q_j||_2^2\right)\right] \\
      & = \arg\max_{q \in \Delta(\mathcal{V})}\left[\nabla f(q_j)^{\mathsf{T}}q - \nabla f(q_j)^{\mathsf{T}} q_j-\frac{1}{2\eta}\left(||q||_{2}^{2} - 2 \langle q, q_j \rangle \right) + \frac{1}{2\eta}||q_j||_2^2\right]. 
 \end{align*}
 Notice from the above that $\nabla f(q_j)^{\mathsf{T}} q_j$ and $||q_j||_2^2$ are independent from the optimisation variable $q$. Removing those and multiplying by $\eta$, we get: 
 \begin{align*}
     q_{j+1} &= \arg\max_{q \in \Delta(\mathcal{V})}\left[\eta\nabla f(q_j)^{\mathsf{T}}q  -\frac{1}{2}||q||_{2}^{2} + \langle q, q_j \rangle \right] \\
     & = \arg\max_{q \in \Delta(\mathcal{V})}\left[\langle q, q_j + \eta \nabla f(q_j)\rangle - \frac{1}{2} || q||_{2}^{2}\right].
 \end{align*}
 If we call $y = q_j + \eta \nabla f(q_j)$, we get the following optimisation problem: 
 \begin{equation*}
     q_{j+1} = \arg\min_{q \in \Delta(\mathcal{V})}\left[\frac{1}{2} ||q||_2^2 - \langle q,y\rangle \right]. 
 \end{equation*}
 Since $y$ is independent of $q$, we can add $\frac{1}{2}||y||_{2}^2$ to our minimisation problem to complete the squares, giving us: 
 \begin{align*}
     q_{j+1} = \arg\min_{q \in \Delta(\mathcal{V})}\left[\frac{1}{2}||q||_2^2 - \langle q, y\rangle + \frac{1}{2}||y||_2^2\right] = \arg\min_{q \in \Delta(\mathcal{V})}\frac{1}{2} \left[||q - y||_2^2\right]. 
 \end{align*}
 From the last line, we see that the projected-gradient update is exactly a Euclidean projection: 
 \begin{equation*}
     q_{j+1} = \prod_{\Delta(\mathcal{V})}(y), \quad \text{where} \quad \prod_{\Delta(\mathcal{V})}(y) = \arg\min_{q \in \Delta(\mathcal{V})}\frac{1}{2}||q - y||_2^{2}.
 \end{equation*}
 In other words, projected gradient ascent produces the next iterate by choosing the distribution $q$ on the simplex that is closest to the unconstrained step $y$ in squared $L_2$ distance. This reveals an important (and often overlooked) fact: projected gradient ascent is not ``geometry neutral'': it is the solution to a proximal subproblem with an $L_2$ regulariser, i.e., it implicitly assumes Euclidean geometry. While this is a reasonable choice in $\mathbb{R}^{n}$, it is a poor match for probability distributions on the simplex, which form a constrained manifold whose natural notion of distance is typically divergence-like (e.g., KL) rather than Euclidean. This mismatch is precisely what mirror descent corrects by replacing the squared $L_2$
 proximity term with a Bregman divergence that respects the simplex geometry.
 \subsection{Bregman Divergences \& Mirror Ascent} 
 If decoding lives on the simplex, then using ${L}_2$ geometry is like doing navigation on the Earth with a flat map: it works locally, but it distorts what ``small'' moves mean. Mirror ascent \cite{beck2003mirror, nemirovskij1983problem, shalev2012online} replaces this distortion with the right geometry. Instead of measuring steps by squared distance, it measures them by a Bregman divergence $\mathcal{D}_{\psi}(q, q_j)$, induced by a convex potential $\psi$ \cite{bregman1967relaxation}. The Bregman divergence is defined as: 
 \begin{equation*}
     \mathcal{D}_{\psi}(q,q_j) = \psi(q) - \psi(q_j) - \langle \nabla \psi(q_j), q - q_j, \rangle
 \end{equation*}
 for a strictly convex and differentiable function $\psi: \Delta(\mathcal{V}) \rightarrow \mathbb{R}$, which is called the distance-
generating (or potential) function. Given $\mathcal{D}_{\psi}(q, q_j)$, we rewrite the problem in \eqref{Eq:PGASurrogate} as a Bregman regularised one: 
\begin{equation}
 \label{Eq:PGABreg}
     q_{j+1} = \arg\max_{q \in \Delta(\mathcal{V})}\left[\langle \nabla f(q_j), q-q_j \rangle - \frac{1}{\eta}\mathcal{D}_{\psi}(q,q_j)\right].
 \end{equation}
 \begin{exampleboxtwo}{The $L_2$ Special Case}
It is interesting to see that if we choose $\psi(q)$ to be the Euclidean norm, i.e., $\psi(q)= \frac{1}{2}||q||_2^2$, we would recover Equation \eqref{Eq:PGASurrogate} as a special case of the generalised problem in Equation \eqref{Eq:PGABreg}. To show that, we begin by understanding the Bregman divergence: 
\begin{equation*}
    \mathcal{D}_{\psi}(q,q_j) = \frac{1}{2}||q||_2^2 - \frac{1}{2}||q_j||_2^2 - \langle q_j, q-q_j\rangle = \frac{1}{2}||q -q_j||_2^2.
\end{equation*}
If we replace this last result into Equation \eqref{Eq:PGABreg}, we exactly obtain Equation \eqref{Eq:PGASurrogate}, which we have shown corresponds to the optimisation problem solved with projected gradient ascent. This goes to say that the problem in Equation \eqref{Eq:PGASurrogate} allows us to use a more general notion of divergences
defined through $\psi$, giving us the ability to consider manifolds (e.g., simplexes) that go beyond
Euclidean spaces. 
\end{exampleboxtwo}
\paragraph{Entropic Distance-Generating Functions over the Simplex.} When optimising over a \emph{simplex} \cite{kivinen1997exponentiated}, the standard choice for the distance-generating function is the entropic potential: $\psi(q) = \sum_{i=1}^{n} q^{(i)} \log q^{(i)}$. In here, the resulting Bregman divergence becomes the Kullback-Leibler (KL) divergence. Therefore, the problem in Equation \eqref{Eq:PGABreg} becomes: 
\begin{equation}
 \label{Eq:PGABregTwo}
     q_{j+1} = \arg\max_{q \in \Delta(\mathcal{V})}\left[\eta\langle \nabla f(q_j), q\rangle - \text{KL}(q||q_j)\right].
 \end{equation}
 Now, let us solve the problem in Equation \eqref{Eq:PGABregTwo} to get the final form of our update. Noticing that we have a constraint $\sum_{i=1}^n q^{(i)} = 1$ and ignoring the positivity constraint \footnote{The constraint has been ignored since the log function naturally enforces positivity.} on $q^{(i)}$, the Lagrangian is defined as: 
\begin{equation*}
    \mathcal{L}(q, \nu) = \eta \sum_{i=1}^{n} \left(\frac{\partial  f(q_j)}{\partial q^{(i)}}\right)q^{(i)} -\sum_{i=1}^{n} q^{(i)}\log \frac{q^{(i)}}{q_j^{(i)}} + \nu \left(1-\sum_{i=1}^{n} q^{(i)}\right), 
\end{equation*}
with $\nu$ being the Lagrange multipliers. Consider the partial derivative of $\mathcal{L}(q,\nu)$ for a single component $q^{(i)}$ and set it to zero: 
\begin{align*}
    \frac{\partial \mathcal{L}}{\partial q^{(i)}} &= \eta \left(\frac{\partial f(q)}{\partial q^{(i)}}\right) - (\log q^{(i)} + 1 - \log q_j^{(i)}) - \nu = 0 \\
    & \implies \log \left(\frac{q^{(i)}}{q_j^{(i)}}\right) = \eta \left(\frac{\partial f(q_j)}{\partial q^{(i)}}\right) - (1 + \nu) \\
    & \implies \frac{q^{(i)}}{q_j^{(i)}} = \exp\left( \eta \left(\frac{\partial f(q_j)}{\partial q^{(i)}}\right)\right)\times \exp\left(-(1+\nu)\right) \\
    & \implies q^{(i)} = Cq_{j}^{(i)}\exp\left( \eta \left(\frac{\partial f(q_j)}{\partial q^{(i)}}\right)\right),
\end{align*}
with $C =  \exp\left(-(1+\nu)\right)$. To find this constant, we use the constraint that the sum $\sum_{i=1}^{n} q^{(i)} =1$: 
\begin{align*}
    C \sum_{i=1}^{n} q_{j}^{(i)}\exp\left( \eta \left(\frac{\partial f(q_j)}{\partial q^{(i)}}\right)\right) = 1 \implies C = \frac{1}{\sum_{l=1}^{n} q_{j}^{(l)}\exp\left( \eta \left(\frac{\partial f(q_j)}{\partial q^{(l)}}\right)\right) }.
\end{align*}
Hence, the overall update for the $i^{th}$ component of $q$ at round $j+1$ amounts to: 
\begin{equation}
\label{Eq:FinalMirror}
    q_{j+1}^{(i)} = \frac{q_{j}^{(i)}\exp\left( \eta \left(\frac{\partial f(q_j)}{\partial q^{(i)}}\right)\right)}{\sum_{l=1}^{n} q_{j}^{(l)}\exp\left( \eta \left(\frac{\partial f(q_j)}{\partial q^{(l)}}\right)\right)}. 
\end{equation}
To write in a vectorised form that is amenable to implementation, we define the following as the vector of partial derivatives: 
\begin{align}
    \bm{g}_j = \left[ \frac{\partial f(q_j)}{\partial q^{(1)}}, \dots,  \frac{\partial f(q_j)}{\partial q^{(n)}}\right]^{\mathsf{T}}.
\end{align}
The numerator of Equation \eqref{Eq:FinalMirror} is thus: 
\begin{equation*}
    \text{Numerator} = q_j \odot \exp(\eta \bm{g}_j), 
\end{equation*}
with $\odot$ being the Hadamard or element-wise product. The denominator is simply the $\text{L}_1$ norm of the numerator: $\text{Denominator} = || q_j \odot \exp(\eta \bm{g}_j)||_1$. Therefore: 
\begin{equation}
\label{Eq:MirrorUpdate}
   \text{\scriptsize\sffamily\color{myblue}\scshape ``Mirror Ascent Update'' \quad}    q_{j+1} = \frac{\text{Numerator}}{\text{Denominator}} = \frac{q_j \odot \exp(\eta \bm{g}_j)}{||q_j \odot \exp(\eta \bm{g}_j)||_1}.
\end{equation}

    In practice, computing $\exp(\eta \frac{\partial f}{\partial q^{(i)}})$ directly can lead to numerical overflow if the gradients are large. To prevent this, we use the Log-Sum-Exp trick by subtracting the maximum value $M = \max(\eta \bm{g}_j)$ from the exponents. Because the sum normalises the update, this shift cancels out mathematically but ensures the computer always handles values $\le 1$.

\subsection{Use Case: Best-of-K Samplers}
We have done a lot of groundwork: we reframed decoding as optimisation on the simplex, re-derived several classical decoders as closed-form special cases, and introduced mirror ascent as a practical solver when closed forms are unavailable. The point of all this was not just unification for its own sake. It was to argue that an optimisation view gives us a principled design language for decoding: samplers are best understood as regularisers (and constraints), and new decoding behaviour should be obtained by writing down the objective we actually want to optimise. To make good on this claim, we now present a concrete use case. We introduce a new regulariser, Best-of-K (BoK), designed for settings where we draw multiple samples and care about coverage of high-quality alternatives, not just the properties of a single draw. BoK drops into our master problem as a plug-in term, and mirror ascent gives an immediate, implementable algorithm.

The motivation is simple: many modern pipelines do not sample once; they sample 
K times (self-consistency, rejection sampling, reranking, verifier-based selection, etc.). In that regime, the decoder is no longer judged by the quality of a single draw, but by what the set of K draws contains. Standard decoders were not designed for this: they often waste budget by repeatedly sampling the same high-probability continuation, while allocating too little mass to plausible alternatives that are individually unlikely but collectively valuable when we have multiple tries. What we want instead is a decoding distribution $q_t^{\star}(\cdot)$ that makes good options show up at least once within K samples, i.e., that explicitly trades off model score against multi-sample coverage. This is exactly the behaviour our framework is built to express: we define a coverage utility for K draws, convert it into a regulariser, and plug it into the master objective to obtain the Best-of-K (BoK) sampler.

\paragraph{The hit probability.} Fix a token $v\in \mathcal{V}$ and suppose we draw $K$ i.i.d.\ samples from a decoding distribution $q(\cdot)$. The probability that we \emph{never} sample $v$ is $(1-q(v))^K$. Therefore, the probability that $v$ appears \emph{at least once} among the $K$ samples is:
\begin{equation}
\mathbf{P}\textrm{r}[\text{$v$ appears at least once in $K$ samples}]
= 1 - (1-q(v))^K.
\label{eq:hit_prob}
\end{equation}
This quantity captures what single-sample decoding objectives ignore: in a multi-sample regime, even moderately small probabilities can become valuable if they increase the chance of seeing a useful alternative at least once.

Summing Equation \eqref{eq:hit_prob} over all tokens yields a notion of total coverage. However, we do not want to reward covering junk tokens equally with high-quality ones, so we introduce nonnegative importance weights $w_t(v)\ge 0$ and define the \emph{weighted $K$-coverage}:
\begin{equation}
\mathcal{U}_{K,t}(q)
:= \sum_{v\in \mathcal{V}} w_t(v)\Big(1-(1-q(v))^K\Big).
\label{eq:UK_def}
\end{equation}
Intuitively, $\mathcal{U}_{K,t}(q)$ is large when: \textit{i)} We allocate mass to many tokens that we care about (large $w_t(v)$), and \textit{ii)} this mass is sufficient for them to be hit at least once within $K$ samples. In practice, $w_t(v)$ can be any nonnegative proxy for ``how much we would like to see token $v$ among $K$ draws'' (e.g., a monotone function of the model score $s_t(v)$, a Top-M indicator, or a softened rank-based weight).

A useful property of the hit probability $1-(1-q(v))^K$ is that it exhibits \emph{diminishing returns}:
its marginal gain decreases as $q(v)$ grows. Indeed:
\[
\frac{\partial}{\partial q(v)}\Big(1-(1-q(v))^K\Big)=K(1-q(v))^{K-1},
\]
which is strictly decreasing in $q(v)$ for $K>1$.
Thus, BoK has an inherent ``anti-collapse'' bias: it is more rewarding to allocate probability mass
to under-covered but valuable tokens than to keep increasing the mass of tokens that are already
likely to appear. This is exactly the multi-sample behaviour we want.

\paragraph{BoK as a regulariser.}
Maximising $\mathcal{U}_{K,t}(q)$ alone would encourage spreading probability mass too widely, including onto implausible tokens.
To keep the decoder anchored to the model, we combine coverage with a KL trust region around the model distribution
$p_t(\cdot)$ and define the BoK regulariser
\begin{equation}
\Omega^{(\mathrm{BoK})}_t(q)
:= \mathrm{KL}(q\|p_t) - \beta\,\mathcal{U}_{K,t}(q),
\label{eq:bok_reg}
\end{equation}
where $\beta\ge 0$ controls how strongly we reward coverage.

Plugging Equation \eqref{eq:bok_reg} into our master problem yields the BoK decoding objective:
\begin{align}
q_t^\star
&= \arg\max_{q\in\Delta(\mathcal{V})}
\Big[
\langle q, s_t\rangle
- \lambda\,\Omega^{(\mathrm{BoK})}_t(q)
\Big]
\nonumber\\
&= \arg\max_{q\in\Delta(\mathcal{V})}
\Big[
\langle q, s_t\rangle
- \lambda\,\mathrm{KL}(q\|p_t)
+ \bar\beta\,\mathcal{U}_{K,t}(q)
\Big],
\label{eq:bok_master}
\end{align}
with $\bar\beta := \lambda\beta$ for convenience.

\paragraph{Mirror-Ascent Update for $\Omega^{(\mathrm{BoK})}$.} Unfortunately, exact closed-forms are hard to come by for $\Omega^{(\mathrm{BoK})}$. Thus, we make use of our mirror ascent update rule in Equation \ref{Eq:MirrorUpdate}. For a single decoding step, BoK-regularised objective is defined as: 
\begin{equation}
\label{Eq:BoKObjective}
f(q)
=
\langle q, s_t\rangle
-\lambda\,\mathrm{KL}(q\|p_t)
+\bar\beta\,\mathcal{U}_{K,t}(q),
\end{equation}
where $q\in\Delta(\mathcal{V})$, $p_t\in\Delta(\mathcal{V})$ is the reference/model distribution, and
\begin{equation}
\label{Eq:BoKUtility}
\mathcal{U}_{K,t}(q)
=
\sum_{v \in \mathcal{V}} w_t(v^{(i)})\Big(1-(1-q^{(i)})^K\Big).
\end{equation}
Define the vector of partial derivatives at iterate $q_j$ as
\begin{equation}
\label{Eq:gjBoK}
\bm g_j
=
\left[
\frac{\partial f(q_j)}{\partial q^{(1)}},
\dots,
\frac{\partial f(q_j)}{\partial q^{(|\mathcal{V}|)}}
\right]^{\mathsf T}.
\end{equation}
For each coordinate $i\in\{1,\dots,n\}$, the partial derivative admits the closed form (we denote $w^{(i)}_t = w_t(v^{(i)})$)
\begin{equation}
\label{Eq:BoKPartial}
\frac{\partial f(q_j)}{\partial q^{(i)}}
=
s_t^{(i)}
-\lambda\left(\log\frac{q_j^{(i)}}{p_t^{(i)}}+1\right)
+\bar\beta\,w_t^{(i)}\,K\,(1-q_j^{(i)})^{K-1}.
\end{equation}
Substituting Equation \eqref{Eq:BoKPartial} into the mirror-ascent update from Equation \eqref{Eq:FinalMirror} yields the explicit BoK update:
\begin{equation}
\label{Eq:FinalMirrorBoK}
q_{j+1}^{(i)}
=
\frac{
q_j^{(i)}\exp\!\Bigg(
\eta\Big[
s_t^{(i)}
-\lambda\Big(\log\frac{q_j^{(i)}}{p_t^{(i)}}+1\Big)
+\bar\beta\,w_t^{(i)}K(1-q_j^{(i)})^{K-1}
\Big]
\Bigg)
}{
\sum_{l=1}^{n}
q_j^{(l)}\exp\!\Bigg(
\eta\Big[
s_t^{(l)}
-\lambda\Big(\log\frac{q_j^{(l)}}{p_t^{(l)}}+1\Big)
+\bar\beta\,w_t^{(l)}K(1-q_j^{(l)})^{K-1}
\Big]
\Bigg)
}.
\end{equation}

Of course, with $\bm g_j$ as in Equation \eqref{Eq:gjBoK}, the update can be written compactly as:
\begin{equation}
\label{Eq:FinalMirrorBoKVec}
q_{j+1}
=
\frac{q_j \odot \exp(\eta \bm g_j)}{\|q_j \odot \exp(\eta \bm g_j)\|_1}.
\end{equation}

Algorithm~\ref{Alg:BoKMirror} summarises how to compute the BoK decoding distribution at a single time step.
We initialise the iterate with the model distribution $q_0=p_t$ (a natural warm-start), then perform $J$ mirror-ascent steps.
Each step computes the vector of partial derivatives $\bm g_j$ using the closed form in Equation \eqref{Eq:BoKPartial}, and applies the
multiplicative update in Equation \eqref{Eq:FinalMirrorBoKVec}, and renormalises to stay on the simplex.
In practice, we implement the update with a Log-Sum-Exp stabilisation by subtracting $M=\max(\eta \bm g_j)$ before exponentiating.

\begin{algorithm}[t]
\caption{BoK Decoder via Mirror Ascent (one decoding step)}
\label{Alg:BoKMirror}
\begin{algorithmic}[1]
\Require scores $s_t\in\mathbb{R}^n$, reference distribution $p_t\in\Delta^n$, weights $w_t\in\mathbb{R}^n_{\ge 0}$
\Require hyperparameters $K\in\mathbb{N}$, $\lambda\ge 0$, $\bar\beta\ge 0$, stepsize $\eta>0$, iterations $J\in\mathbb{N}$
\State Initialise $q_0 \leftarrow p_t$
\For{$j=0,1,\dots,J-1$}
    \For{$i=1,2,\dots,n$}
        \State $g_j^{(i)} \leftarrow s_t^{(i)}-\lambda\left(\log\frac{q_j^{(i)}}{p_t^{(i)}}+1\right)+\bar\beta\,w_t^{(i)}K(1-q_j^{(i)})^{K-1}$
    \EndFor
    \State $M \leftarrow \max(\eta \bm g_j)$ \Comment{Log-Sum-Exp stabilisation}
    \State $\tilde q_{j+1} \leftarrow q_j \odot \exp(\eta \bm g_j - M\mathbf{1})$
    \State $q_{j+1} \leftarrow \tilde q_{j+1} / \|\tilde q_{j+1}\|_1$
\EndFor
\State \textbf{return} $q_J$
\end{algorithmic}
\end{algorithm}

\subsection{Use Case Evaluation: How good are BoK Samplers?}\label{Sec:BoK_Eval}
We evaluate the BoK Sampler as a practical, decoding-time regulariser for improving multi-sample generation without additional training or external verifiers. The evaluation is designed to answer three questions: (i) whether BoK improves solution quality over standard sampling baselines; (ii) whether the gains are robust across decoding temperatures and different hyperparameter choices in the underlying optimisation; and (iii) what compute overhead is incurred when solving the BoK objective with mirror ascent. Our main finding is that BoK consistently improves or matches standard decoding across tasks and models, with the largest gains appearing in higher-temperature regimes where vanilla sampling is more diverse but less reliable. BoK leverages the regularised objective to retain diversity while increasing the chance of sampling high-quality alternatives with only a small computational overhead.

\begin{table}[t]
\centering
\small
\setlength{\tabcolsep}{6pt}
\renewcommand{\arraystretch}{1.15}
\resizebox{0.85\linewidth}{!}{%
\begin{tabular}{lccccc}
\toprule
\textbf{Method} & \textbf{$\tau$=0.10} & \textbf{$\tau$=0.25} & \textbf{$\tau$=0.50} & \textbf{$\tau$=0.70} & \textbf{$\tau$=0.90} \\
\midrule
\multicolumn{6}{l}{\textbf{MATH500 (Qwen2.5-Math-7B)}} \\
\hspace{1.5em}Base      & 72.2 & 72.6 & 69.4 & 64.4 & 53.0 \\
\hspace{1.5em}Top-K     & 72.8 & \textbf{73.4} & 69.4 & 65.0 & 56.2 \\
\hspace{1.5em}BoK (Ours) $\beta{=}0.01,\ \lambda{=}0.1$ & \textbf{74.2} & 72.8 & 71.0 & \textbf{73.0} & 71.2 \\
\hspace{1.5em}BoK (Ours) $\beta{=}0.02,\ \lambda{=}0.2$ & 72.6 & 71.8 & 72.2 & 72.4 & \textbf{71.6} \\
\hspace{1.5em}BoK (Ours) $\beta{=}0.05,\ \lambda{=}0.5$ & 72.8 & 71.6 & \textbf{72.8} & 72.8 & 70.8 \\
\midrule
\multicolumn{6}{l}{\textbf{GPQA (Qwen2.5-Math-7B)}} \\
\hspace{1.5em}Base      & 32.32 & 33.84 & 26.26 & 31.31 & 30.30 \\
\hspace{1.5em}Top-K     & \textbf{33.84} & 35.35 & 28.79 & 31.31 & 31.82 \\
\hspace{1.5em}BoK (Ours) $\beta{=}0.01,\ \lambda{=}0.1$ & 30.81 & 33.84 & 31.82 & 33.33 & \textbf{36.36} \\
\hspace{1.5em}BoK (Ours) $\beta{=}0.02,\ \lambda{=}0.2$ & 31.31 & \textbf{36.36} & 33.33 & \textbf{34.85} & 31.82 \\
\hspace{1.5em}BoK (Ours) $\beta{=}0.05,\ \lambda{=}0.5$ & \textbf{33.84} & 32.83 & \textbf{35.35} & 31.82 & 31.82 \\
\midrule
\multicolumn{6}{l}{\textbf{HumanEval (Qwen2.5-Math-7B)}} \\
\hspace{1.5em}Base      & \textbf{56.71} & 53.05 & 48.78 & 49.39 & 32.93 \\
\hspace{1.5em}Top-K     & 54.88 & 51.83 & \textbf{51.83} & 46.34 & 37.80 \\
\hspace{1.5em}BoK (Ours) $\beta{=}0.01,\ \lambda{=}0.1$ & 54.27 & \textbf{54.88} & 50.0 & \textbf{54.27} & \textbf{52.44} \\
\hspace{1.5em}BoK (Ours) $\beta{=}0.02,\ \lambda{=}0.2$ & 54.88 & \textbf{54.88} & 51.83 & 53.05 & \textbf{52.44} \\
\hspace{1.5em}BoK (Ours) $\beta{=}0.05,\ \lambda{=}0.5$ & 56.01 & 51.83 & \textbf{51.83} & 52.44 & 51.22 \\
\bottomrule
\end{tabular}%
}
\caption{Accuracy across temperatures for Qwen2.5-Math-7B on MATH500, GPQA, and HumanEval. We report Base, Top-K (K=50), and BoK with three representative $(\beta, \lambda)$ settings, illustrating robustness across both temperature and hyperparameters.}
\label{tab:bok_temp_qwen25math7b}
\end{table}

\begin{table}[t]
\centering
\small
\setlength{\tabcolsep}{6pt}
\renewcommand{\arraystretch}{1.15}
\resizebox{0.8\linewidth}{!}{%
\begin{tabular}{lccccc}
\toprule
\textbf{Method} & \textbf{$\tau$=0.10} & \textbf{$\tau$=0.25} & \textbf{$\tau$=0.50} & \textbf{$\tau$=0.70} & \textbf{$\tau$=0.90} \\
\midrule
\multicolumn{6}{l}{\textbf{MATH500 (Qwen2.5-7B)}} \\
\hspace{1.5em}Base      & 57.6 & 60.4 & 56.6 & 50.2 & 44.2 \\
\hspace{1.5em}Top-K     & \textbf{59.1} & 59.4 & 52.8 & 51.6 & 41.0 \\
\hspace{1.5em}BoK (Ours) $\beta{=}0.01,\ \lambda{=}0.1$ & 57.6 & 59.6 & 60.6 & 60.0 & 58.0 \\
\hspace{1.5em}BoK (Ours) $\beta{=}0.02,\ \lambda{=}0.2$ & 58.8 & 59.2 & \textbf{61.8} & \textbf{60.4} & \textbf{60.2} \\
\hspace{1.5em}BoK (Ours) $\beta{=}0.05,\ \lambda{=}0.5$ & \textbf{59.1} & \textbf{61.0} & 59.4 & 59.4 & 59.4 \\
\midrule
\multicolumn{6}{l}{\textbf{GPQA (Qwen2.5-7B)}} \\
\hspace{1.5em}Base      & \textbf{32.83} & 27.78 & 21.72 & 25.76 & 24.24 \\
\hspace{1.5em}Top-K     & 28.28 & 29.80 & 27.78 & 29.80 & \textbf{32.32} \\
\hspace{1.5em}BoK (Ours) $\beta{=}0.01,\ \lambda{=}0.1$ & 31.82 & 29.80 & 29.80 & 29.29 & \textbf{32.32} \\
\hspace{1.5em}BoK (Ours) $\beta{=}0.02,\ \lambda{=}0.2$ & 31.82 & 29.29 & \textbf{30.30} & 27.78 & 29.29 \\
\hspace{1.5em}BoK (Ours) $\beta{=}0.05,\ \lambda{=}0.5$ & 30.30 & \textbf{30.30} & 29.80 & \textbf{29.29} & 24.24 \\
\midrule
\multicolumn{6}{l}{\textbf{HumanEval (Qwen2.5-7B)}} \\
\hspace{1.5em}Base      & 70.13 & 67.68 & 71.34 & 71.95 & 45.12 \\
\hspace{1.5em}Top-K     & 71.95 & 70.73 & 71.95 & 65.24 & 57.93 \\
\hspace{1.5em}BoK (Ours) $\beta{=}0.01,\ \lambda{=}0.1$ & \textbf{72.56} & 71.95 & 71.95 & \textbf{72.56} & \textbf{71.34} \\
\hspace{1.5em}BoK (Ours) $\beta{=}0.02,\ \lambda{=}0.2$ & 71.59 & \textbf{73.17} & \textbf{74.78} & 71.51 & 69.51 \\
\hspace{1.5em}BoK (Ours) $\beta{=}0.05,\ \lambda{=}0.5$ & \textbf{72.56} & 72.56 & 69.51 & 69.51 & 66.46 \\
\bottomrule
\end{tabular}%
}
\caption{Accuracy across temperatures for Qwen2.5-7B on MATH500, GPQA, and HumanEval (same baselines and BoK settings as Table~\ref{tab:bok_temp_qwen25math7b}).}
\label{tab:bok_temp_qwen257b}
\end{table}

\paragraph{Experimental setup.} We evaluate BoK on a math-specialised model \texttt{Qwen2.5-Math-7B} and a general-purpose model \texttt{Qwen2.5-7B}, across three complementary benchmarks spanning math, QA, and code: \texttt{MATH500}~\cite{lightman2023let}, \texttt{GPQA-diamond}~\cite{rein2024gpqa}, and \texttt{HumanEval}~\cite{chen2021evaluating}. We compare BoK against standard autoregressive sampling (Base) from the model distribution at temperature $\tau$, and Top-K, sampling restricted to the top-$K$ tokens per step with renormalisation (we fix $K{=}50$ across tasks and temperatures). All methods use the same prompts (Qwen default prompts~\cite{yang2024qwen2}) and evaluation scripts, the same maximum generation length $T_{\max}{=}3072$ with early stopping on \texttt{EOS}, and the same random-seed protocol.

\paragraph{Results and Analysis.} We implement the BoK sampler following Algorithm~\ref{Alg:BoKMirror}, using mirror-ascent updates to compute a per-step sampling distribution $q_t^{\star}(\cdot)$ that increases exploration via the coverage utility while remaining anchored to the base model through KL regularisation (Eq.~\eqref{eq:bok_master}). In particular, $\beta$ controls the strength of the coverage reward, and $\lambda$ controls the strength of KL anchoring toward the reference distribution $p_t$. Across tasks and models, BoK matches or improves upon Base and Top-K sampling in most settings, as shown in Tables~\ref{tab:bok_temp_qwen25math7b} and~\ref{tab:bok_temp_qwen257b}.

\noindent The gains are most pronounced at higher temperatures, where vanilla sampling becomes more diverse but less reliable. For \texttt{Qwen2.5-Math-7B} on \texttt{MATH500} at $\tau{=}0.9$, BoK increases accuracy from 53.0\% to 71.6\% (+18.6\%) and exceeds Top-K at 56.2\% by +15.4\%. We observe a similar pattern on \texttt{GPQA} and \texttt{HumanEval}, where BoK improves over the baselines at $\tau{=}0.9$ by +6.06\% and +14.64\%. Similar improvements hold for \texttt{Qwen2.5-7B}, where BoK substantially mitigates the accuracy drop at high temperatures. At lower temperatures, improvements are naturally smaller; in a few near-deterministic regimes, Base or Top-K sampling can remain slightly stronger or comparable to BoK, which is consistent with the reduced need for exploration when the model distribution is already sharply peaked.

\noindent The results also indicate that BoK is not overly sensitive to hyperparameter choice: multiple $(\beta,\lambda)$ settings yield competitive performance across temperatures, with the best pair varying by task and regime. Overall, the consistent improvements across all three tested configurations suggest a stable operating region. BoK thus provides a practical and robust way to trade off diversity and reliability at decoding time without exhaustive hyperparameter search, with particularly strong benefits in high-entropy sampling settings.

\begin{table}[t]
\centering
\small
\setlength{\tabcolsep}{10pt}
\renewcommand{\arraystretch}{1.15}
\resizebox{0.8\linewidth}{!}{%
\begin{tabular}{lcccccc}
\toprule
\textbf{Gradient steps} & Base & 2 & 5 & 10 & 15 & 20 \\
\midrule
\textbf{MATH500 acc. (\%)} & 64.4 &  69.6 & 73.0 & 71.6 & 71.2 & 72.8 \\
\textbf{Runtime (s)} & 15.84 & 15.87 & 16.88 & 17.70 & 17.91  & 18.26 \\
\bottomrule
\end{tabular}%
}
\caption{Effect of the number of mirror-gradient steps per token on MATH500 accuracy for Qwen2.5-MATH-7B (BoK) using $
\tau=0.7$, $\beta=0.01$ and $\lambda=0.1$.}
\label{tab:bok_steps_math500_qwen257b}
\end{table}

\paragraph{Efficiency and practical deployment.} In practice, BoK introduces only a small runtime overhead because computing the BoK-optimal distribution $q_t^\star$ requires only a few mirror-ascent updates per token. All BoK results in Tables~\ref{tab:bok_temp_qwen25math7b} and~\ref{tab:bok_temp_qwen257b} use $5$ mirror-ascent steps per token. Across the three benchmarks, the additional cost remains modest relative to base decoding: on MATH500, BoK runs in 16.88s vs.\ 15.84s; on GPQA, 17.60s vs.\ 15.43s; and on HumanEval, BoK is slightly faster in our implementation (8.65s vs.\ 9.74s), which we attribute to shorter generations under BoK. This suggests that the mirror-ascent solver converges quickly, so only a small number of optimisation steps are required to obtain an effective $q_t^\star$. 

\noindent To make this explicit, we further study the effect of the number of mirror-gradient steps per token on the MATH500 task, as shown in Table~\ref{tab:bok_steps_math500_qwen257b}. There is an improvement compared with the base decoding, even with a small number of steps: using only 2 steps improves accuracy from 64.4\% to 69.6\% with a negligible runtime increase (15.87s), and 5 steps reach 73.0\% with a still modest overhead (16.88s). Increasing the number of steps beyond 5 yields only marginal changes in accuracy while steadily increasing runtime. Overall, these results indicate that BoK remains effective with a small, fixed number of mirror-ascent steps, making it suitable for practical decoding-time use.

\section{Conclusion and Future Work}
We argued that decoding should not be viewed as a collection of disconnected heuristics.
Instead, many widely used decoding rules arise as exact solutions to a single optimisation template on the probability simplex,
where different decoding behaviours correspond to different regularisers and feasible sets.
This perspective unifies classical procedures (e.g., greedy, Softmax, Top-K/Top-P, and sparse decoders) under one ``Master'' problem,
and makes explicit the role of optimality conditions in determining which tokens become active or inactive.
When closed-form solutions are unavailable, we showed that mirror ascent provides a principled, simplex-native solver whose updates preserve
valid distributions by construction.
Finally, to demonstrate that the framework is not merely explanatory, we introduced a concrete use case, the Best-of-K (BoK) regulariser, which
directly targets multi-sample coverage and can be implemented with the same mirror-ascent machinery.

The optimisation view suggests several promising directions.
First, while we focused on per-step decoding, it would be natural to extend the framework to \emph{sequence-level} objectives that couple decisions across time \cite{ji2026scalablepowersamplingunlocking, karan2025reasoningsamplingbasemodel},
e.g., enforcing coverage, length, or style constraints globally rather than locally. Second, BoK illustrates how multi-sample utilities can be encoded as regularisers; a broader family of \emph{compute-aware} objectives could be explored,
including utilities that model downstream reranking, verifier selection, or self-consistency more explicitly \cite{aggarwal2023letssamplestepstep, chen2023universalselfconsistencylargelanguage, wang2023selfconsistencyimproveschainthought, zimmer2025rethinkinglargelanguagemodel}.
Third, mirror ascent opens the door to richer constraint sets beyond the simplex, such as structured sparsity, group constraints, or dynamic support sets that depend on
external tools or retrieval modules \cite{brown2025systematicliteraturereviewretrievalaugmented, lewis2021retrievalaugmentedgenerationknowledgeintensivenlp, schick2023toolformerlanguagemodelsteach, shi2025toollearningwildempowering}.
We hope this work helps shift decoder design from folklore to first-principles objective design.
\vspace{0.55em}
\begin{tcolorbox}[
  enhanced,
  boxrule=0pt,
  colback=black!1,
  borderline west={2pt}{0pt}{black!15},
  arc=2mm,
  left=3mm,right=3mm,top=1.0mm,bottom=1.0mm,
  width=\linewidth
]
\centering
{\fontsize{11.2}{13.2}\selectfont\sffamily\bfseries\color{black!90}
\raisebox{0.05ex}{\large \twemoji{pushpin}}\hspace{0.35em}
In short: \textbf{Decoding is not a hack}; it is optimisation!
}
\end{tcolorbox}
 \newpage
\printbibliography

\end{document}